\documentclass[runningheads]{llncs}
\usepackage{graphicx}
\usepackage{amsmath,amssymb} 

\usepackage{color}
\usepackage[table]{xcolor}

\usepackage[hidelinks]{hyperref}
\usepackage[capitalize]{cleveref}
\newcommand{\etal}{\textit{et al. }}
\usepackage{caption}
\usepackage{subcaption}

\definecolor{bicycle}{RGB}{0, 128, 0}
\definecolor{bus}{RGB}{0, 128, 128}
\definecolor{car}{RGB}{128, 128, 128}
\definecolor{cow}{RGB}{64, 128, 0}
\definecolor{horse}{RGB}{192, 0, 128}
\definecolor{motorbike}{RGB}{64, 128, 128}
\definecolor{person}{RGB}{192, 128, 128}
\definecolor{sheep}{RGB}{128, 64, 0}
\definecolor{train}{RGB}{128, 192, 0}

\Crefname{section}{Sect.}{Sects.}
\begin{document}
\pagestyle{headings}
\mainmatter

\title{Causes of Catastrophic Forgetting in Class-Incremental Semantic Segmentation} 

\titlerunning{Causes of Catastrophic Forgetting in CiSS}
\authorrunning{T. Kalb and J. Beyerer}

\author{Tobias Kalb\inst{1} \and
		J\"urgen Beyerer\inst{2,3}}
\institute{Porsche Engineering Group GmbH, Porschestraße 911, 71287 Weissach, Germany 
		\email{tobias.kalb@porsche-engineering.de} \and
		Fraunhofer Institute of Optronics, Systems Technologies and Image Exploitation IOSB \and
		Karlsruhe Institute of Technology, 76131 Karlsruhe, Germany\\
		\email{juergen.beyerer@iosb.fraunhofer.de} }

\maketitle

\begin{abstract}
Class-incremental learning for semantic segmentation (CiSS) is presently a highly researched field which aims at updating a semantic segmentation model by sequentially learning new semantic classes. A major challenge in CiSS is overcoming the effects of catastrophic forgetting, which describes the sudden drop of accuracy on previously learned classes after the model is trained on a new set of classes. 
Despite latest advances in mitigating catastrophic forgetting, the underlying causes of forgetting specifically in CiSS are not well understood.
Therefore, in a set of experiments and representational analyses, we demonstrate that the semantic shift of the background class and a bias towards new classes are the major causes of forgetting in CiSS. Furthermore, we show that both causes mostly manifest themselves in deeper classification layers of the network, while the early layers of the model are not affected. Finally, we demonstrate how both causes are effectively mitigated utilizing the information contained in the background, with the help of knowledge distillation and an unbiased cross-entropy loss.
\end{abstract}

\section{Introduction}
Semantic segmentation is a long-standing problem in computer vision, which aims to assign a semantic label to each pixel of an image. 
However, a fundamental constraint of the traditional semantic segmentation benchmarks is that they assume that all classes are known beforehand and are all learned at once. This assumption limits the use of semantic segmentation for practical applications, as within a realistic scenario the model should be able to learn new classes without requiring a complete retraining of all previous ones. For this reason, the recently emerged field of class-incremental semantic segmentation (CiSS) focuses on achieving this goal of incrementally learning new classes.
The two main challenges that CiSS has to overcome are catastrophic forgetting \cite{French1999-FRECFI,MCCLOSKEY1989109} of old classes and the semantic shift of the background class. Recently, several methods were proposed to mitigate the limitations brought on by these challenges \cite{Tasar2019,Michieli2019,klingner2020class,Cermelli2020,Douillard2020,Michieli2021}. Contrary to methods in class-incremental image classification, CiSS methods are mostly based on the concept of knowledge distillation \cite{kd_hinton}. While significant progress has been made in mitigating the effects of catastrophic forgetting and the semantic shift of the background class, there is limited understanding of how the drop in accuracy manifests itself within the CNN-based semantic segmentation model. The focus of our work is to identify the causes of forgetting in CiSS. Specifically, we aim at revealing how activation drift, inter-task-confusion and task-recency bias affect the performance in CiSS and how existing approaches overcome these effects. Furthermore, we present a set of experiments and tools, which can help to gain insights into the internal representations of the models in \cref{measure_methods}.
Using these methods we reveal:
\begin{enumerate}
    \item The semantic shift of the background class is the main cause of the catastrophic drop in performance in CiSS. However, the re-appearance of previous classes in the background in subsequent training tasks in fact reduces the internal activation drift. Forgetting mainly happens in the classification layers of the model, as discriminating features for old classes are not forgotten in the encoder, but are either assigned to visually similar classes or to the background class in the classification layers.
    \item The semantic shift of the background class can be addressed with the unbiased cross entropy loss (UNCE) \cite{Cermelli2020}, which takes the uncertainty of the content of the background into account.
    \item The performance on the test set indicates that forgetting for prior regularization methods is much more severe as it actually is, as the regularization terms effectively mitigate the internal activation drift in the encoder of the segmentation model and only the classification layer is biased towards the background and old classes.
    \item Methods that do not use any form of replay do not learn discriminating features for all classes. Specifically, the model is not able to distinguish old classes from new classes that are visually closely related, e.g. the model is not able to distinguish the classes \textit{train} and \textit{bus} when they are not learned in the same learning step.
\end{enumerate}

\section{Related Work}

\subsection{Continual Learning}
The majority of research in continual learning is focused on developing methods to overcome the effects of catastrophic forgetting. This is achieved by dedicating a subset of parameters to each task either dynamically or explicitly, by penalizing updates on important parameters for the previous task during training on a new task \cite{Aljundi2018_MAS,Kirkpatrick2015,aljundi_survey,synaptic_intelligence}, directly storing data from previous task for later replay \cite{shin2017continual,NEURIPS2019_fa7cdfad,rebuffi-cvpr2017,hayes2020remind}, as well as knowledge-distillation-based methods, which utilize the activations of the old network as regularization term during training on new data \cite{Li2018}.
In a recent empirical survey by Masana \etal \cite{masana2020class} the best performing methods on class-incremental image classification all utilize replay to cope with forgetting \cite{masana2020class}. Other surveys and investigations confirm the notion that class-incremental learning requires some form of replay \cite{VandeVen2020} and that prior regularization methods might fail at learning discriminating features in the class-incremental setting \cite{lesort2020regularization,Hsu2018}. For a state-of-the-art overview on class-incremental image classification we refer to \cite{masana2020class}.

\subsection{Class-Incremental Semantic Segmentation}
When comparing the most successful approaches in class-incremental image classification \cite{masana2020class} and CiSS \cite{Tasar2019,Michieli2019,klingner2020class,Cermelli2020,Douillard2020,Michieli2021,Kalb2021}, it is noticeable that contrary to classification in CiSS most of the recent methods build on the idea of Learning without Forgetting (LwF) \cite{Li2018} and utilize a knowledge distillation-based loss.
In most CiSS benchmarks knowledge distillation-based approaches even outperform replay-based methods \cite{Kalb2021}.
As a result, state-of-the-art CiSS approaches mostly rely on some form of knowledge distillation \cite{kd_hinton} and do not require any form of replay. These approaches have been proven to be effective on their respective benchmarks.
However, they require that previously learned classes will reappear in the future training images, because otherwise they would still suffer from forgetting, as we will confirm in our experiments. Therefore, recent approaches also investigated how to integrate exemplar replay into CiSS \cite{Maracani_2021_ICCV,Douillard2021,Kalb2021}. 
\subsection{Studying effects of Catastrophic Forgetting}
Prior work on understanding catastrophic forgetting in deep learning for image classification used representational analysis techniques such as centered kernel alignment \cite{Kornblith2019} and linear probing to conclude that deeper layers are disproportionately the cause of forgetting in CNNs \cite{davari2021probing,ramasesh2021anatomy}. We will confirm this is also true for semantic segmentation. Other work analyzed the loss-landscape of continually trained models \cite{mirzadeh2021linear} or investigated how the task sequence \cite{Nguyen2019} and task similarity \cite{ramasesh2021anatomy} impact forgetting. Contrary to prior work, we concentrate on identifying the causes of forgetting that arise specifically in CiSS, as the semantic shift of background class and the multi-class nature of semantic segmentation introduce new challenges into the continual learning setting that are not yet well understood.

\section{Problem Formulation}
Before going into details about the causes of forgetting, we define the general task of CiSS.
The goal of semantic segmentation is to assign a class out of a set of pre-defined classes $\mathcal{C}$ to each pixel in a given image. 
A training task $T = \{(x_m, y_m)\}^{M}_{m=1}$ consists of a set of $M$ images $x\in \mathcal{X}$ with $\mathcal{X} = \mathbb{R}^{H \times W \times 3}$ and corresponding labels $y\in \mathcal{Y}$ with $\mathcal{Y} = \mathcal{C}^{H \times W}$. 
Given the task $T$ the goal in semantic segmentation is to learn a mapping $f_{\theta} : \mathcal{X} \to \mathbb{R}^{H\times W\times |\mathcal{C}|}$ from the image space $\mathcal{X}$ to a posterior probability vector $q$.
The output segmentation mask for a single pixel $i$ of the image is obtained as $\bar{y}_i = \operatorname{arg\,max}_{c\in \mathcal{C}}q_{i, c}$.
In the class-incremental learning setting the model $f_\theta$ is not trained on a single task $T$ but on a sequence of tasks $T_t$. Each task $T_t$ extends the previous set of classes $\mathcal{C}_{t-1}$ by a set of novel classes $\mathcal{S}_t$ resulting in the new label set $\mathcal{C}_t = \mathcal{C}_{t-1} \cup \mathcal{S}_t$. In this setting, the labels of classes $\mathcal{C}_{t-1}$ are not included in the training set of $T_t$, with the exception of the background class $b$. After learning a task $T_t$, the model is required to correctly discriminate between all the observed classes $\mathcal{C}_t$.

\subsection{Causes of Forgetting in Class-Incremental Learning}\label{sec:causes}
\begin{figure}[]
  \centering
  \includegraphics[width=0.50\textwidth]{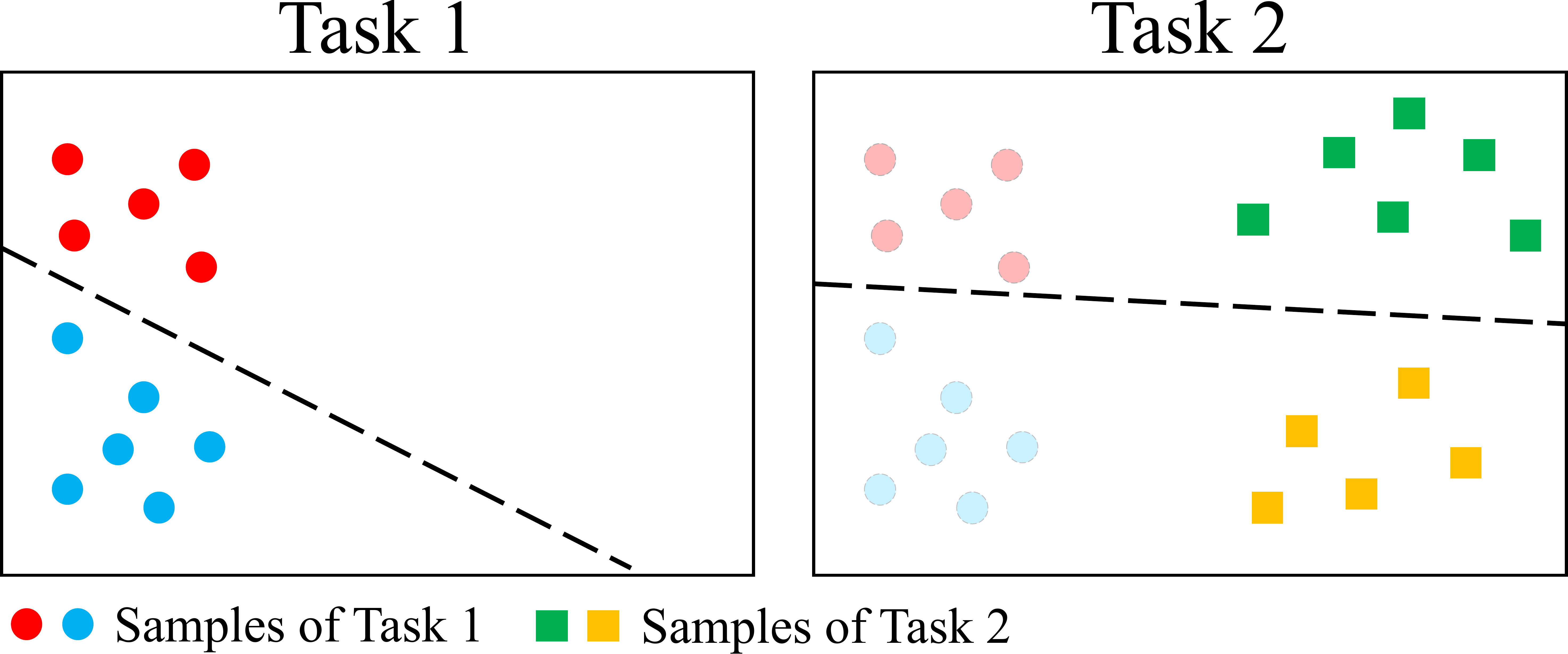}
  \caption{Visualization of task confusion in class-incremental learning. As classes of Task 1 (circles) and classes of Task 2 (squares) are never trained at the same time, the classifier never learns to discriminate between circles and squares, which causes inter-task confusion.}
  \label{fig:inter-task-conf}
\end{figure}

The fundamental challenge in class-incremental learning is that during optimization of the model on a new $T_t$, the model is optimized without regard to the previous classes $\mathcal{C}_{t-1}$, which leads to catastrophic forgetting of previous classes.
Masana \etal \cite{masana2020class} stated four causes of forgetting in class-incremental classification. 
\begin{itemize}
    \item \textbf{Weight Drift:} During optimization on task $T_t$, the weights of the model that were relevant to the previous task $T_{t-1}$ are updated without regard to the previous task, resulting in drop of performance on task $T_{t-1}$.
    \item \textbf{Activation Drift:} A change of the weights of the model directly results in a change of internal activations and to the output of the model. 
    \item \textbf{Inter-task confusion:} The objective in class-incremental learning is to correctly discriminate between between all the observed classes $\mathcal{C}_t$. However, as the classes are never jointly trained, the learned features are not optimized to discriminate classes from different tasks, as shown in \cref{fig:inter-task-conf}.
    \item \textbf{Task-recency bias:} In the class-incremental setting, the model is optimized to predict new classes without regarding the old classes. This leads to a strong bias for the most recently learned classes.
    This bias can easily be seen in the confusion matrix, as shown in \cref{fig:conf_FT}. Furthermore, especially in the final classification layer, the weights and biases of the classifier layer have higher magnitudes of the weights vectors for new classes \cite{Hou_2019_CVPR}. In CiSS with the addition of the background class, an additional bias to the background class can be observed. 
\end{itemize}
In the following we want to investigate how these causes manifest themselves in CiSS and how existing approaches mitigate these causes.

\section{Methods to measure forgetting}\label{measure_methods}
Purely accuracy based evaluation only allows restrictive insight into the causes of forgetting of a model. Therefore, in this section, we present the key methods that we use in our analysis, including layer matching \cite{Csiszarik2021} and decoder retraining accuracy. We utilize these methods to measure the activation shift between a model $f_0$ and $f_1$. The model $f_0$ is trained on $T_0$, whereas $f_1$ is initialized with the parameters of $f_0$ and incrementally trained on $T_1$.

\subsection{Layer Matching with Dr. Frankenstein}\label{layer_stitch}
\begin{figure}[]
  \centering
  \includegraphics[width=\textwidth]{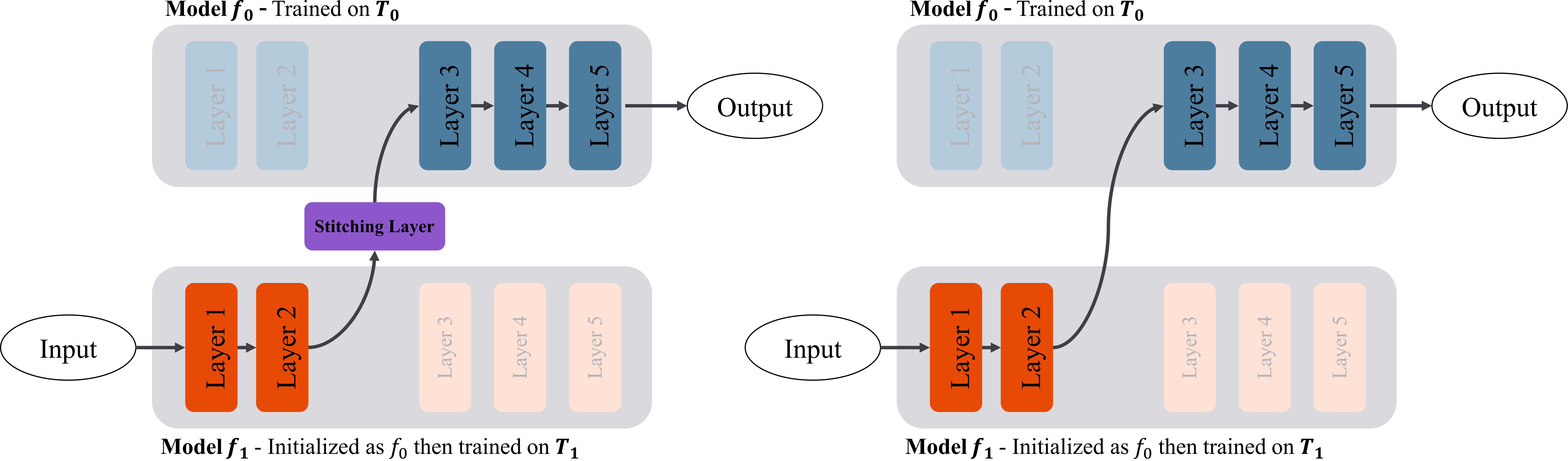}
  \caption{Comparing a) the original Dr. Frankenstein layer matching approach of using an additional stitching layer as proposed by \cite{Csiszarik2021} and b) our approach for measuring activation drift in continual learning by directly propagating the activations of $f_1$ to $f_0$ without a stitching layer.}
  \label{fig:layer_stitch}
\end{figure}

The Dr. Frankenstein toolset proposed by Csisz\'{a}rik \etal \cite{Csiszarik2021} aims to match the activations of two neural networks on a given layer by joining them with a stitching layer, compare \cref{fig:layer_stitch}. The goal of the stitching layer is to transform the activations of a specific layer of $f_0$ to the corresponding activations of a model $f_1$.
The stitching layer is initialized using least-squares-matching and is optimized using the loss function the network was trained with. In order to measure the similarity of the learned representations, the accuracy of the resulting Frankenstein Network is evaluated on the test set and compared to the initial accuracy of the model $f_0$. The higher the resulting relative accuracy is, the closer the learned representations of the models are to each other.
In our analysis we omit the stitching layer and directly use the activations of $f_0$ in $f_1$, as we noticed in our experiments that the initial activations are already very similar. We attribute this to the fact that the models are closely related because $f_1$ is initialized with the parameters of $f_0$. Our setup can bee seen on the right side in \cref{layer_stitch}. 
We denote the resulting Frankenstein Network when layer $n$ of model $f_1$ is stitched to layer $n+1$ of model $f_0$ as $f_{1,0}^{n}$, i.e. the network depicted in \cref{layer_stitch} would be $f_{1, 0}^{2}$.
If the accuracy of the resulting Frankenstein network is not affected, this is clear evidence that the internal representations of $f_1$ were not altered drastically during training on $T_1$. This analysis will give insight into how much the activation at a specific layer has changed after training continually, but will give no insight into a possible positive backward transfer.

\subsection{Decoder Retrain Accuracy}\label{decoder_retrain}
Inspired by use of linear probing for continual learning \cite{davari2021probing}, which measures representational forgetting by calculating the difference in accuracy an optimal linear classifier achieves on an old task before and after introducing a new task, we propose Decoder Retraining Accuracy. To measure the Decoder Retraining Accuracy, we freeze the encoder of the model and retrain the decoder on all classes with the same training configuration and measure the mIoU on a test set. Instead of measuring the representational drift as in \cite{davari2021probing} this will give a measure of how useful the learned representations are to discriminate between classes of different tasks, effectively measuring the contribution of inter-task-confusion of the encoder.

\section{Experiments}
\paragraph{Datasets:}
We conduct our experiments using the PascalVoc-2012 \cite{pascal-voc-2012} dataset, which contains 20 object classes and a background class. We follow the established CiSS PascalVoc-15-5 split that is widely used \cite{Douillard2020,Michieli2019,Michieli2021,Cermelli2020}. The PascalVOC-15-5 split is a two step incremental learning task, which consists of learning 15 classes (1--15) in the first step $T_0$ and the remaining 5 classes (16--20) in the second step $T_1$.\footnote{As the focus of this paper is to understand the general causes of forgetting in CiSS, we leave the study of the impact of different splits, more classes and longer task sequences to future work.}
We follow the two distinct class-incremental settings for \textit{Disjoint} and \textit{Overlapped} proposed by Cermelli \etal \cite{Cermelli2020}. 
In both settings, only the set of current classes $\mathcal{S}_t$ is labelled, while the rest is labelled as background $b$. However, in the \textit{Disjoint} setting, the images of the current task $T_t$ only contain pixels of classes $\mathcal{C}_t$, meaning that images that contain pixels belonging to classes of future tasks will be discarded in the training set of $T_t$. In the \textit{Overlapped} setting, pixels can belong to any of the classes, but classes that do not belong to current training set will be labeled as background. Finally, in order to study the impact of the semantic shift that the background class is subjected to in the \textit{Disjoint} and \textit{Overlapped}, we introduce the \textit{Fully Disjoint} setting. In this setting each task only contains pixels belonging to the current set of classes, which therefore avoids the interference that originates from the semantic shift of the background class. We utilize this setting to study the impact of the semantic background shift on catastrophic forgetting, but we note that this is an unrealistic scenario, as in semantic segmentation classes naturally re-appear.

\paragraph{Model:} Similar to \cite{Kalb2021} we use ERFNet \cite{erfnet} in our evaluation, as the underlying effects of forgetting are similar to more established models like DeepLabV3 \cite{Chen2017}, while at the same time ERFNet is more susceptible to forgetting due to its smaller size, which exaggerates the effect of the causes of forgetting. We do not use any pre-trained models in our experiments, as pre-training is known to increase robustness to catastrophic forgetting \cite{Gallardo2021,Mehta2021}. Instead, we use the same randomly initialized weights for every method in our experiments.

\paragraph{Methods Compared:}
In our evaluation, we focus on evaluating naive fine-tuning approaches, representative regularization and replay methods as these cover the basic concepts that are considered as potential solutions for CiSS. For prior regularization methods we consider EWC \cite{Kirkpatrick2015} and MAS \cite{Aljundi2018_MAS}. For data regularization methods we use LwF \cite{Li2018}, with the modification of \cite{klingner2020class}, in which the distillation loss is only applied to the parts of the image that are labelled as background. We explicitly do not consider incremental improvements of knowledge distillation based approaches of \cite{Douillard2020,Cermelli2020,Michieli2021,Maracani_2021_ICCV}, as our focus is on evaluating the underlying causes of forgetting. For replay we save 20 samples for each class in the buffer. For every experiment we also list results for the \textit{offline} model, which is jointly trained on all classes in one step. Further information regarding the implementation details and selected hyperparameters can be found in the supplementary material.

\subsection{Semantic Background Shift and Class Confusion}\label{general_results}
\begin{figure}[]
    \centering
    \begin{minipage}[t]{0.7\textwidth}
    \begin{subfigure}[t]{0.3\textwidth}
        \includegraphics[width=\textwidth]{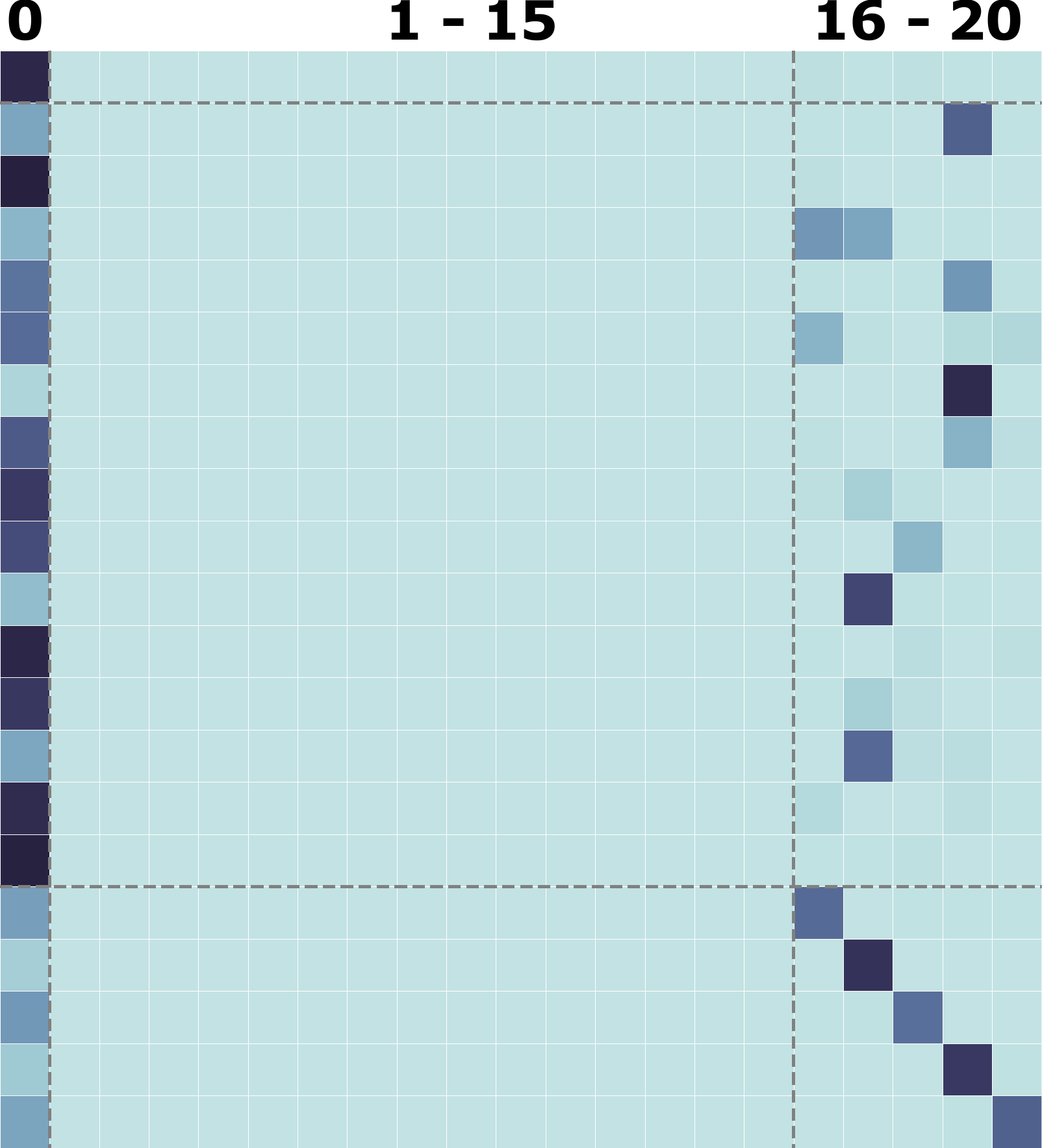}
        \caption{Fine-Tuning}
        \label{fig:conf_FT}
    \end{subfigure}
    \begin{subfigure}[t]{0.3\textwidth}
        \includegraphics[width=\textwidth]{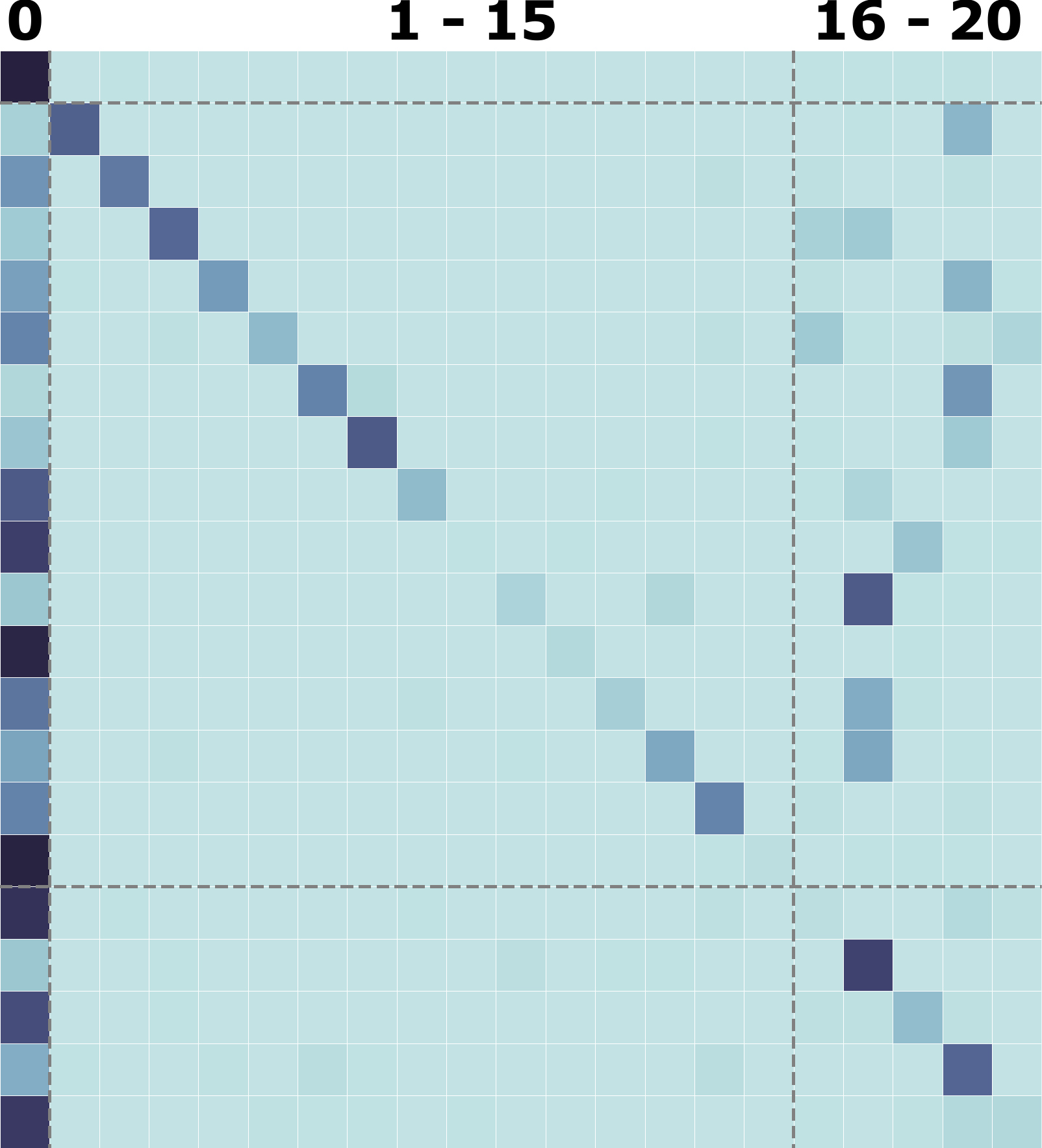}
        \caption{MAS}
        \label{fig:conf_MAS}
    \end{subfigure}
    \begin{subfigure}[t]{0.3\textwidth}
        \includegraphics[width=\textwidth]{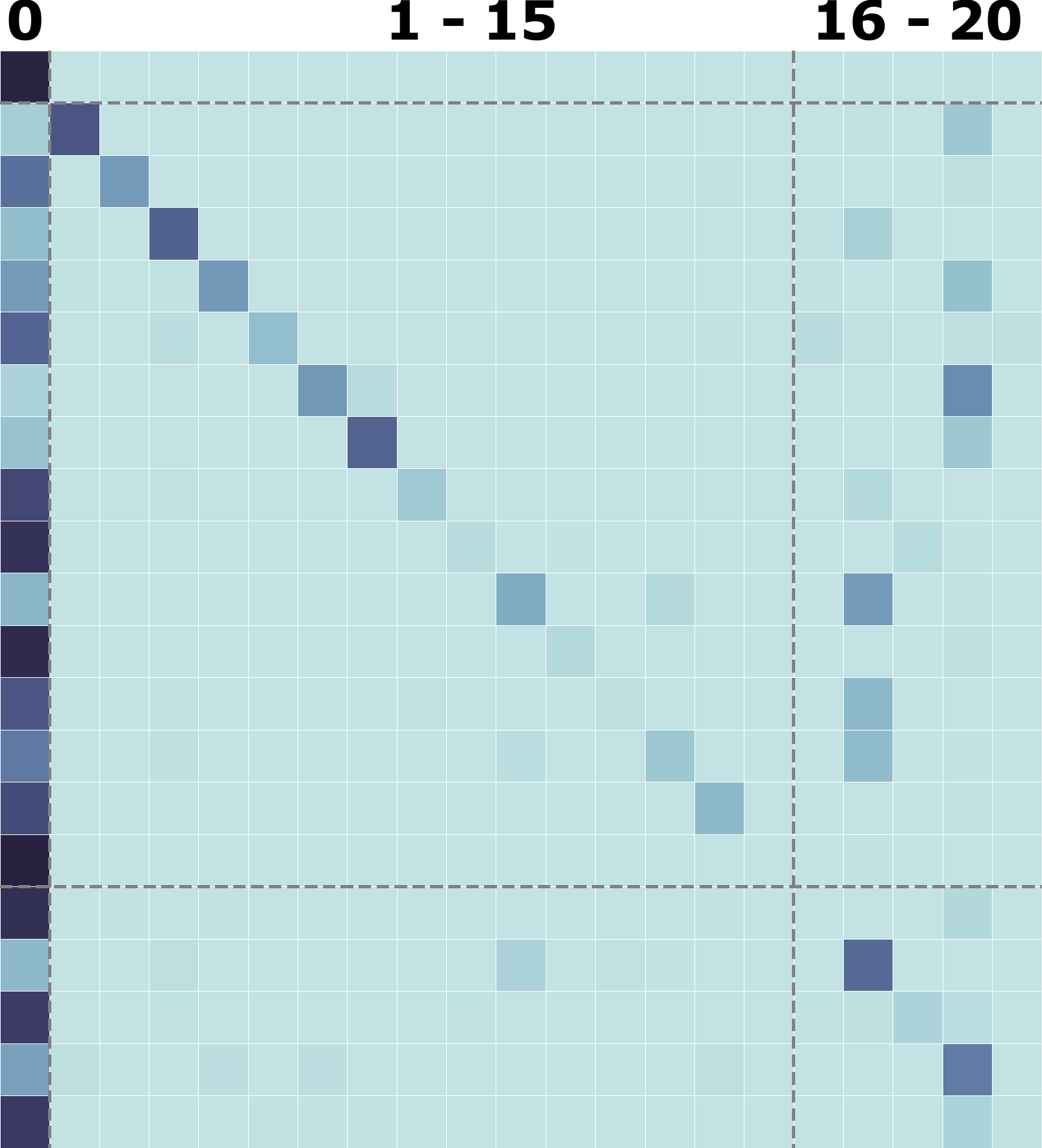}
        \caption{EWC}
        \label{fig:conf_EWC}
    \end{subfigure}
    
    \begin{subfigure}[t]{0.3\textwidth}
        \includegraphics[width=\textwidth]{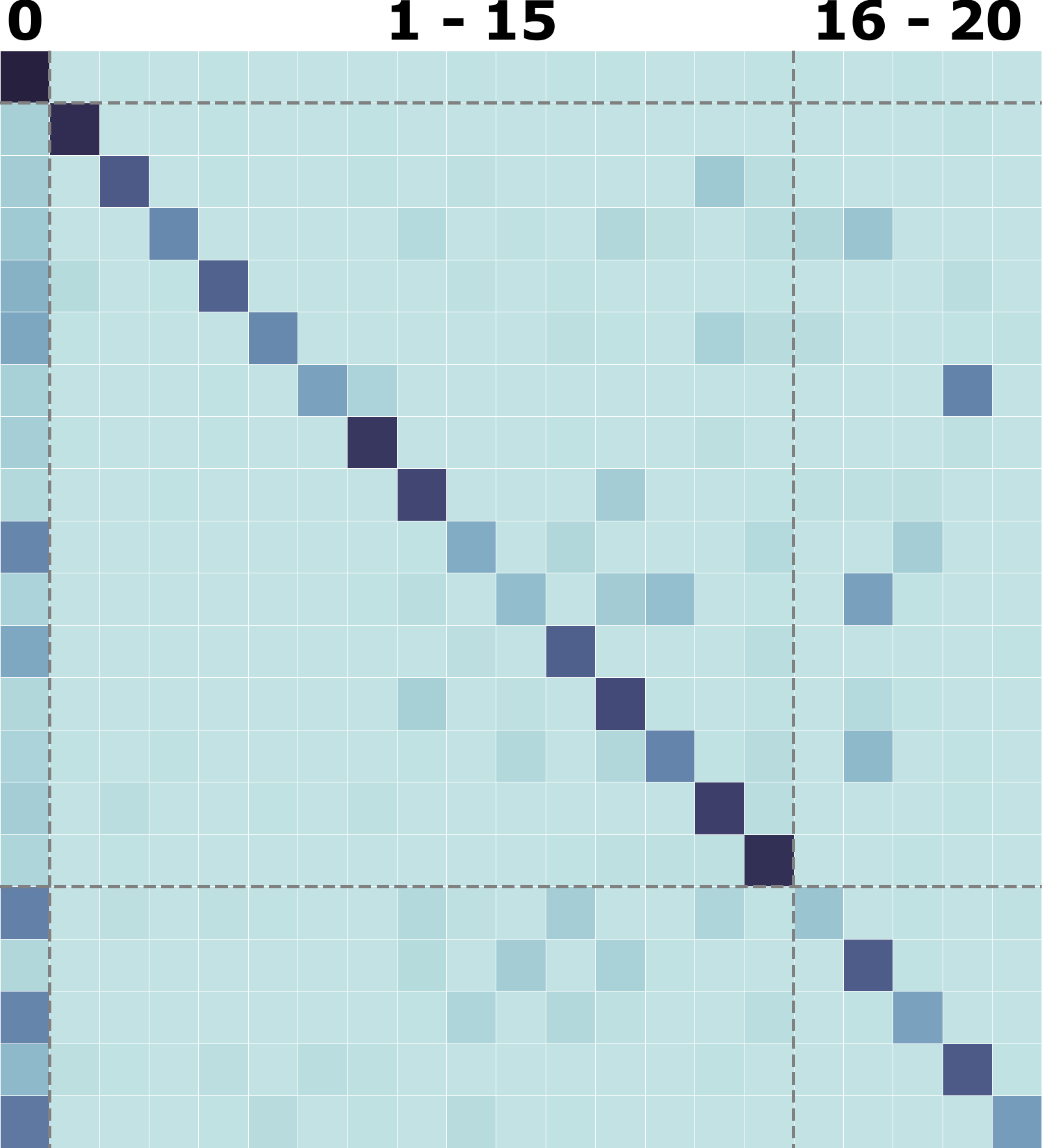}
        \caption{LwF}
        \label{fig:conf_LWF}
    \end{subfigure}
    \begin{subfigure}[t]{0.3\textwidth}
        \includegraphics[width=\textwidth]{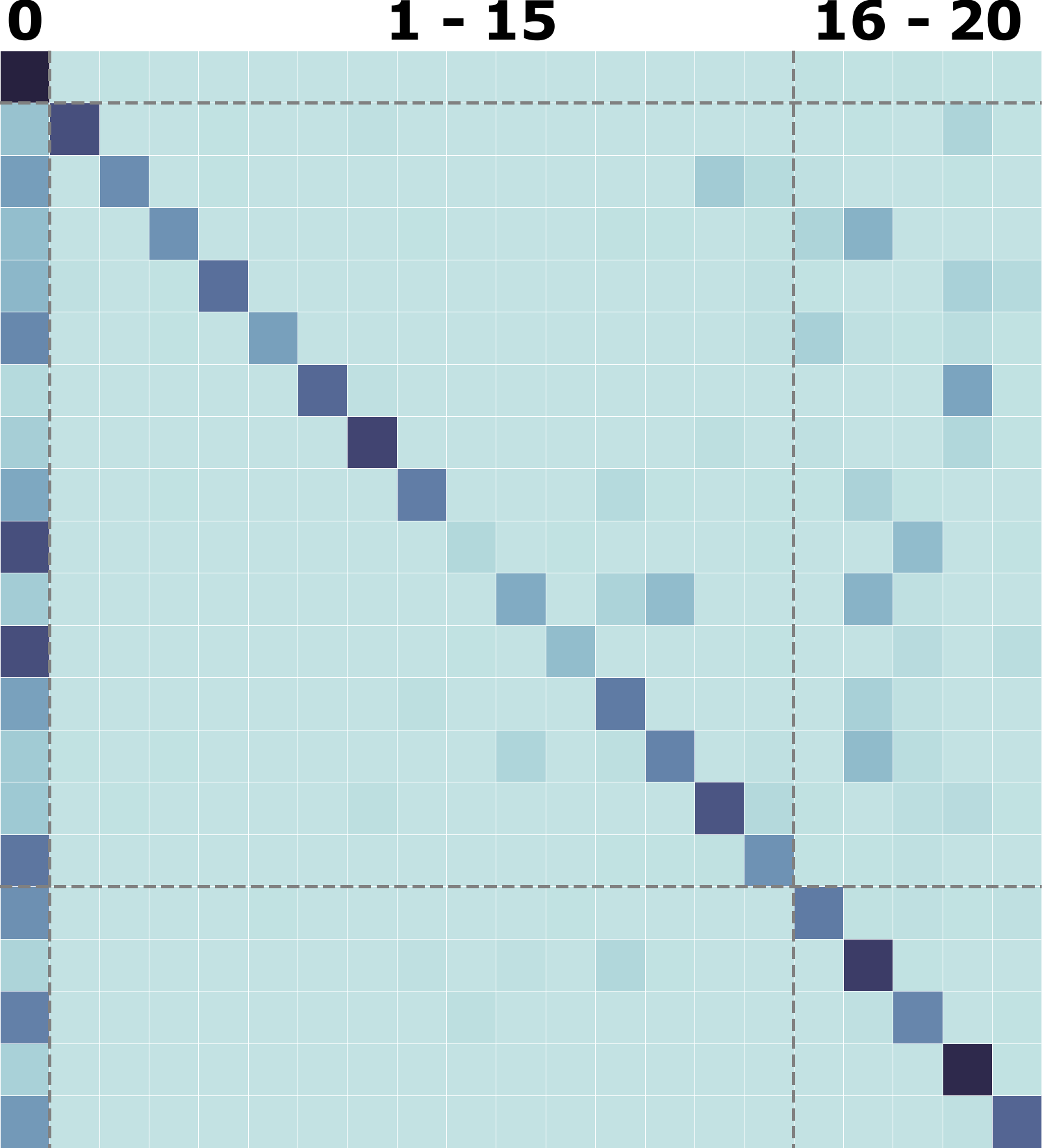}
        \caption{Replay}
        \label{fig:conf_Replay}
    \end{subfigure}
        \begin{subfigure}[t]{0.3\textwidth}
        \includegraphics[width=\textwidth]{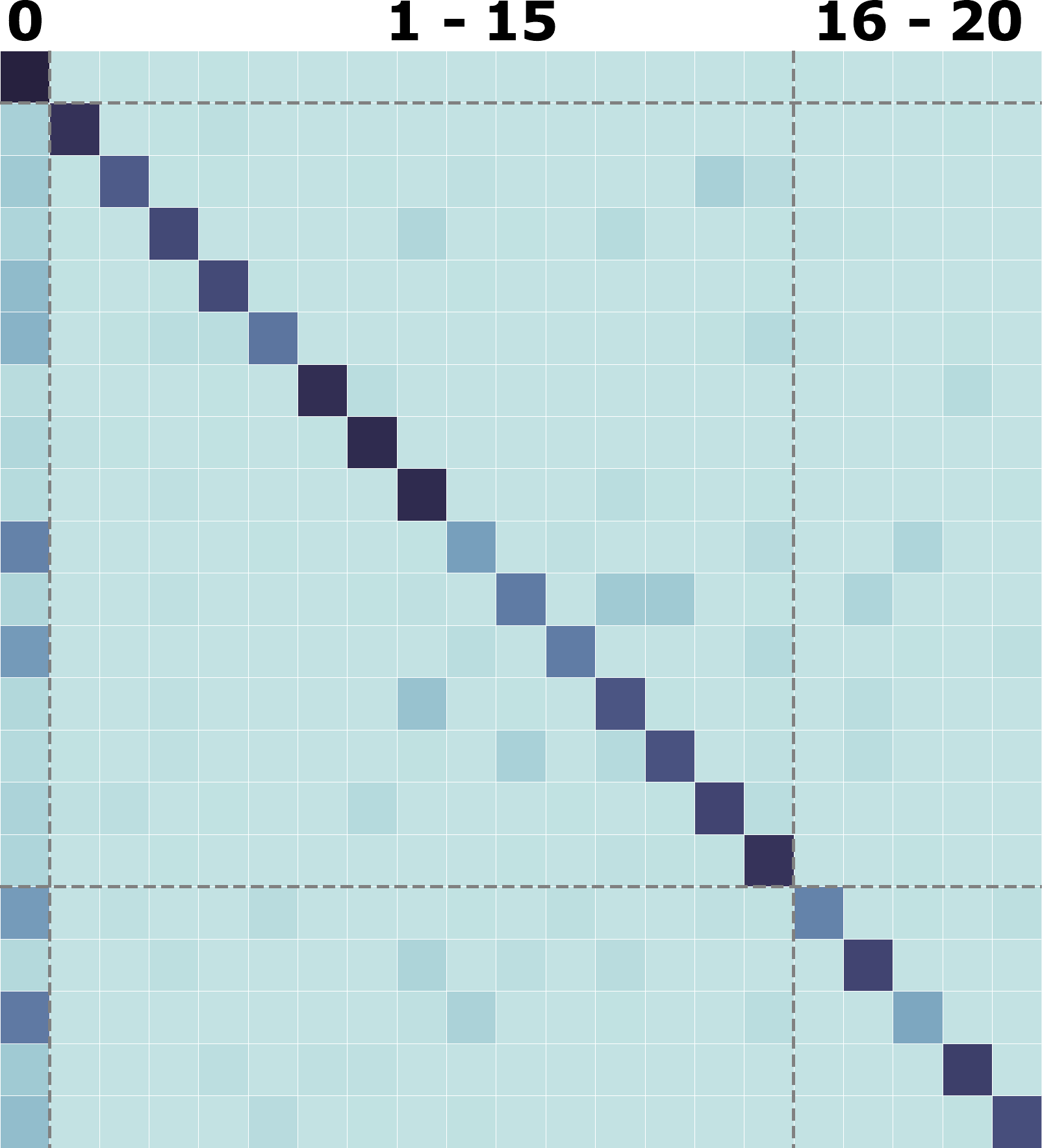}
        \caption{Offline}
        \label{fig:conf_MASUNCE}
    \end{subfigure}
    \end{minipage}
    \begin{minipage}[t]{0.125\textwidth}
        \centering
        \resizebox{\textwidth}{!}{%
        \begin{tabular}{l|l}
        \textbf{ID} & \textbf{Name} \\ \hline
        0 & background \\
        1 & aeroplane \\
        2 & bicycle \\
        3 & bird \\
        4 & boat \\
        5 & bottle \\
        6 & bus \\
        7 & car \\
        8 & cat \\
        9 & chair \\
        10 & cow \\
        11 & dining table \\
        12 & dog \\
        13 & horse \\
        14 & motorbike \\
        15 & person \\
        16 & potted plant \\
        17 & sheep \\
        18 & sofa \\
        19 & train \\
        20 & monitor
        \end{tabular}%
        }
    \end{minipage}
    \caption{Confusion matrices after training on PascalVoc-15-5 (disjoint). The confusion matrix for Fine-Tuning a) shows a severe bias to the background class and the classes of the most recent task (16-20). EWC \cite{Kirkpatrick2015} and MAS \cite{Aljundi2018_MAS} decrease the bias in exchange for worse accuracy on the most recent classes. 
    Replay and LwF \cite{Li2018} reduce the bias towards new classes and the background.}
\end{figure}

First, we study the impact of the semantic background shift on forgetting by comparing the results of the selected CiSS methods on the \textit{Overlapped}, \textit{Disjoint}, and \textit{Full Disjoint} tasks. These tasks have a varying degree of semantic shift of the background class. The results are displayed in \cref{tab:overall_results}. For the \textit{Overlapped} and \textit{Disjoint} tasks, it can be noted that only LwF and Replay effectively learn to discriminate between all classes. EWC and MAS effectively mitigate forgetting of old classes (0--15) compared to Fine-Tuning, but they also inhibit the learning of new classes (16--20). The reason for the low mIoU for EWC and MAS on all classes can be inferred from the confusion matrix, in which EWC and MAS exhibit a strong bias to the background class and a minor bias towards a few new classes, shown in in \cref{fig:conf_MAS,fig:conf_EWC}. LwF and Replay reduce both biases. In the \textit{Overlapped} and \textit{Disjoint} setting, LwF and Replay achieve similar performance, as in this setting LwF can effectively replay old classes by discovering them in the background of new images. 
However, once these classes do not re-appear in the background, as is the case in the \textit{Full Disjoint} task, LwF develops a strong bias towards selected new classes. In contrast to this, Replay benefits from the \textit{Full Disjoint} setting as the training is no longer affected by the semantic background shift.
Similarly, MAS and EWC also show significant improvement in this setting, as they benefit from the fact that old classes do not appear as background in the new task, thus not interfering with previously learned knowledge. This is especially noticeable in the confusion matrices of the \textit{Full Disjoint} setting, in which the bias towards the background class is greatly reduced\footnote{The confusion matrix are shown in the supplementary material.}.
This demonstrates that the semantic shift of the background class is a significant cause of forgetting for prior regularization methods like EWC and MAS and that it has a noticeable effect on replay as well. The results also indicate that knowledge distillation is the most effective method to combat this semantic shift.\\
Finally, upon a closer look at the semantics of the false positives, we see that old classes that are falsely assigned to a new class share a semantic and visual relationship. In this case \textit{bus} (6), \textit{car} (7), \textit{boat} (5) are assigned to \textit{train} (19), whereas \textit{cow} (10) and \textit{horse} (13) are classified as \textit{sheep} (17). The remaining classes that do not share such a relationship with new classes are falsely classified as background. This confusion can only be alleviated by either Replay or LwF when old classes re-appear in the background in subsequent tasks.
To clarify which layers of the model contribute the most to the bias for the background class and newly learned classes, next up we investigate the internal activation drift.

\begin{table}[]
	\centering
	\caption{Results of Semantic Segmentation on Pascal-VOC 2012 in Mean IoU (\%) on the overlapped, disjoint and full disjoint settings.}
	\label{tab:overall_results}
	\resizebox{0.8\textwidth}{!}{%
		\begin{tabular}{lccccccccc}
			\multicolumn{10}{c}{\textbf{PascalVoc 15-5 Semantic Segmentation}} \\ \hline
			\multicolumn{1}{l|}{} & \multicolumn{3}{c|}{\textbf{Overlapped}} & \multicolumn{3}{c|}{\textbf{Disjoint}} & \multicolumn{3}{c}{\textbf{Full Disjoint}} \\
	
			\multicolumn{1}{l|}{\textbf{Method}} & \textit{0-15} & \multicolumn{1}{c|}{\textit{16-20}} & \multicolumn{1}{c|}{\textit{all}} & \textit{0-15} & \multicolumn{1}{c|}{\textit{16-20}} & \multicolumn{1}{c|}{\textit{all}} & \textit{0-15} & \multicolumn{1}{c|}{\textit{16-20}} & \textit{all} \\ \hline
			
            \multicolumn{1}{l|}{Fine-Tuning} & 4.5 & \multicolumn{1}{c|}{22.2} & \multicolumn{1}{c|}{8.8}   & 4.6 & \multicolumn{1}{c|}{23.0} & \multicolumn{1}{c|}{9.0}   & 5.1& \multicolumn{1}{c|}{16.3} & 7.8 \\ 
            \multicolumn{1}{l|}{MAS \cite{Aljundi2018_MAS}} & 24.1 & \multicolumn{1}{c|}{10.8} & \multicolumn{1}{c|}{21.0}   & 30.6 & \multicolumn{1}{c|}{12.9} & \multicolumn{1}{c|}{26.4}   & 35.6& \multicolumn{1}{c|}{12.9} & 30.2 \\ 
            \multicolumn{1}{l|}{EWC \cite{Kirkpatrick2015}} & 23.8 & \multicolumn{1}{c|}{11.8} & \multicolumn{1}{c|}{21.0}   & 28.1 & \multicolumn{1}{c|}{10.1} & \multicolumn{1}{c|}{23.8}   & 35.2& \multicolumn{1}{c|}{10.9} & 29.4 \\ 
            \multicolumn{1}{l|}{Replay} & 41.3 & \multicolumn{1}{c|}{31.5} & \multicolumn{1}{c|}{39.0}   & 42.2 & \multicolumn{1}{c|}{29.1} & \multicolumn{1}{c|}{39.1}   & 48.2& \multicolumn{1}{c|}{28.8} & 43.6 \\ 
            \multicolumn{1}{l|}{LwF \cite{Li2018}} & 45.8 & \multicolumn{1}{c|}{28.2} & \multicolumn{1}{c|}{41.6}   & 44.4 & \multicolumn{1}{c|}{25.4} & \multicolumn{1}{c|}{39.9}   & 35.9& \multicolumn{1}{c|}{12.6} & 30.4 \\ 
    
			\hline
			\multicolumn{1}{l|}{Offline} & 55.7 & \multicolumn{1}{c|}{47.6} & \multicolumn{1}{c|}{53.8}  &  55.7 & \multicolumn{1}{c|}{47.6} & \multicolumn{1}{c|}{53.8}  &55.7 & \multicolumn{1}{c|}{47.6} & 53.8
	\end{tabular}}
\end{table}

\subsection{Measuring the effect of Activation Drift}\label{activation_drift}
\begin{figure}[]
\resizebox{\textwidth}{!}{%
\begin{tabular}{ccccccc}
 \textbf{Input} & \textbf{Ground Truth} & \textbf{LwF} &  \textbf{MAS}  &  \textbf{MAS $f_{1,0}^{17}$} & \textbf{MAS $f_{1,0}^{18}$} & \textbf{MAS+UNCE} \\
 \includegraphics{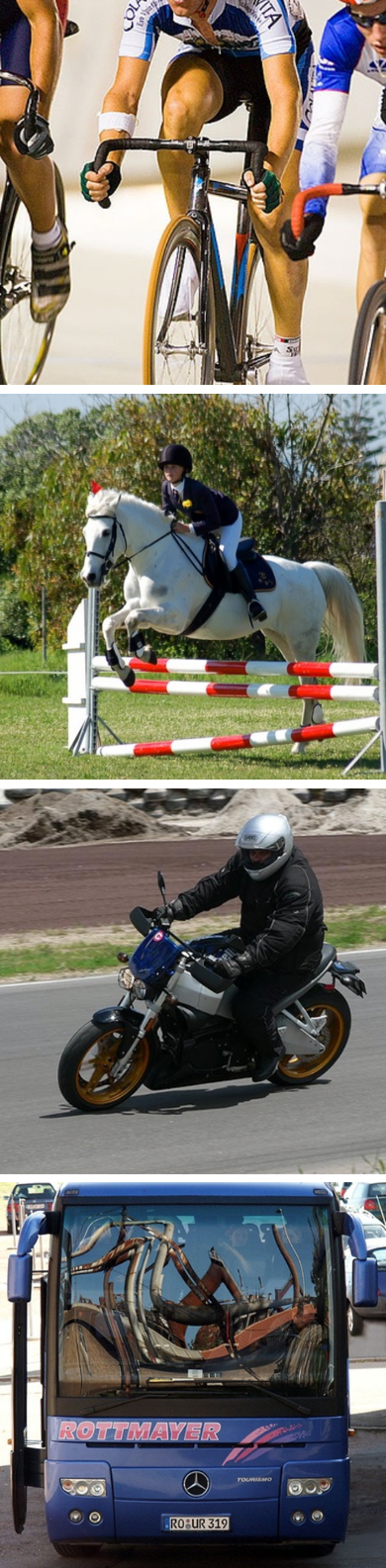} &  
 \includegraphics{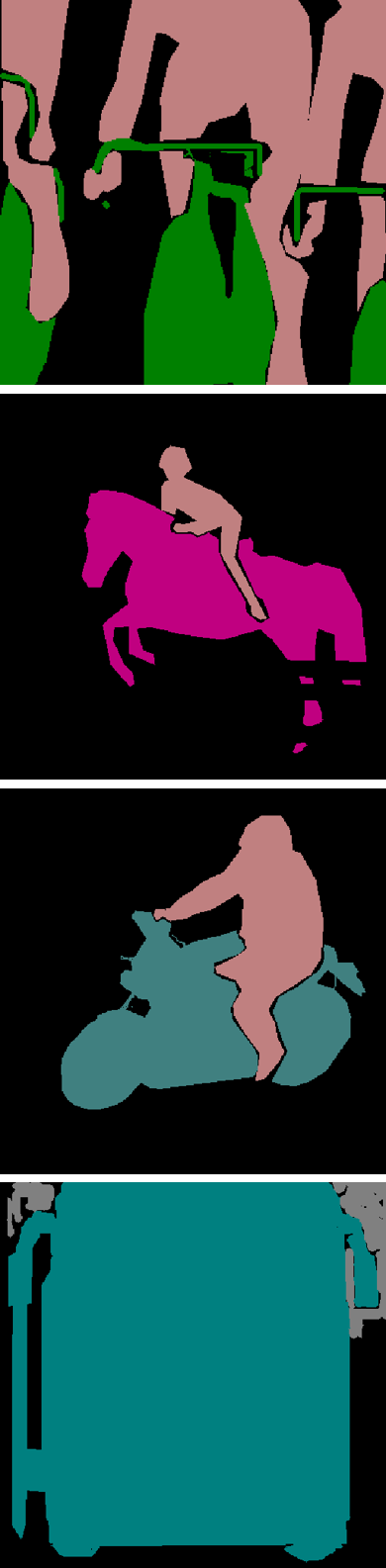} &  
 \includegraphics{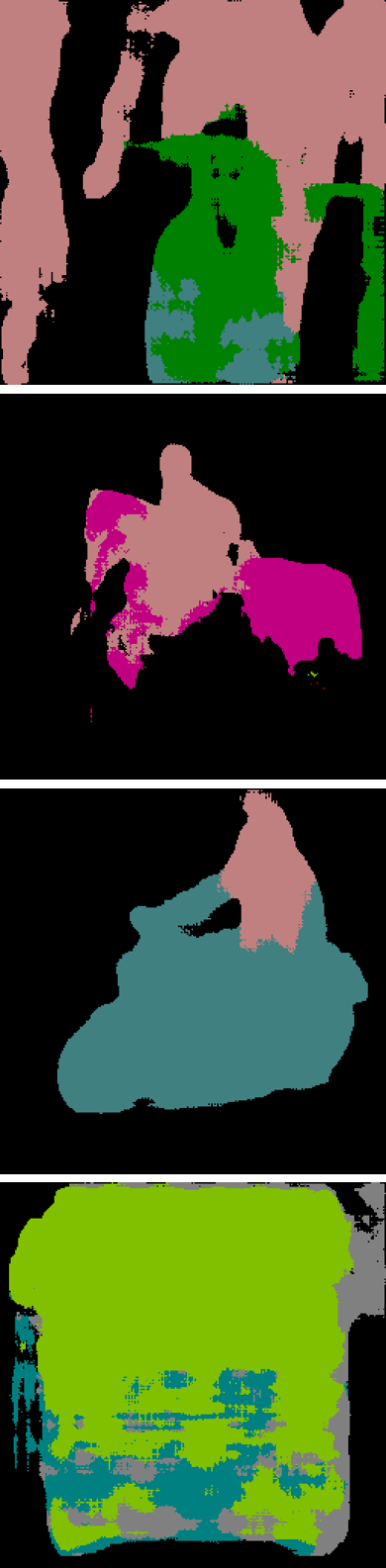} &  
 \includegraphics{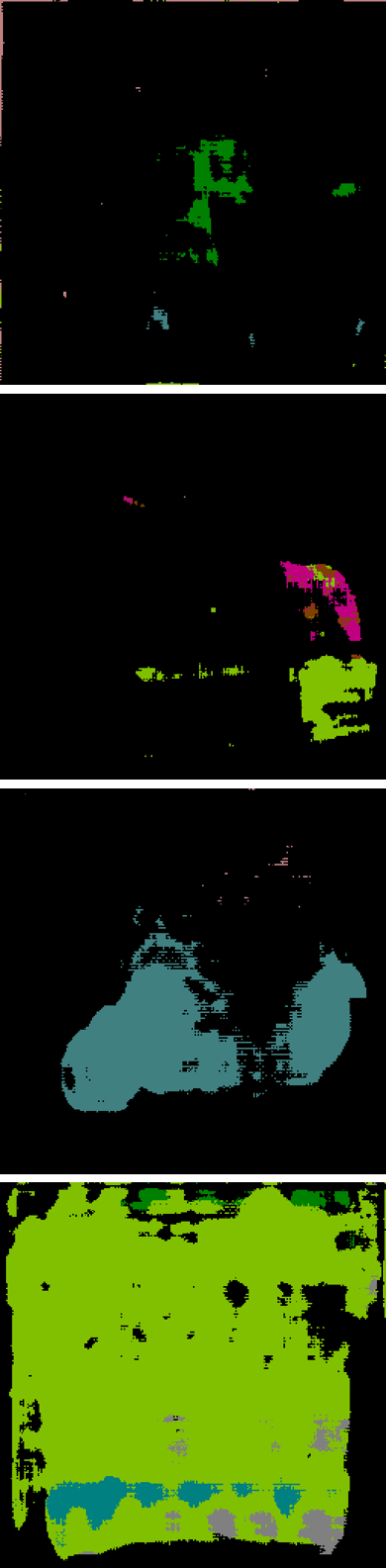}&  
 \includegraphics{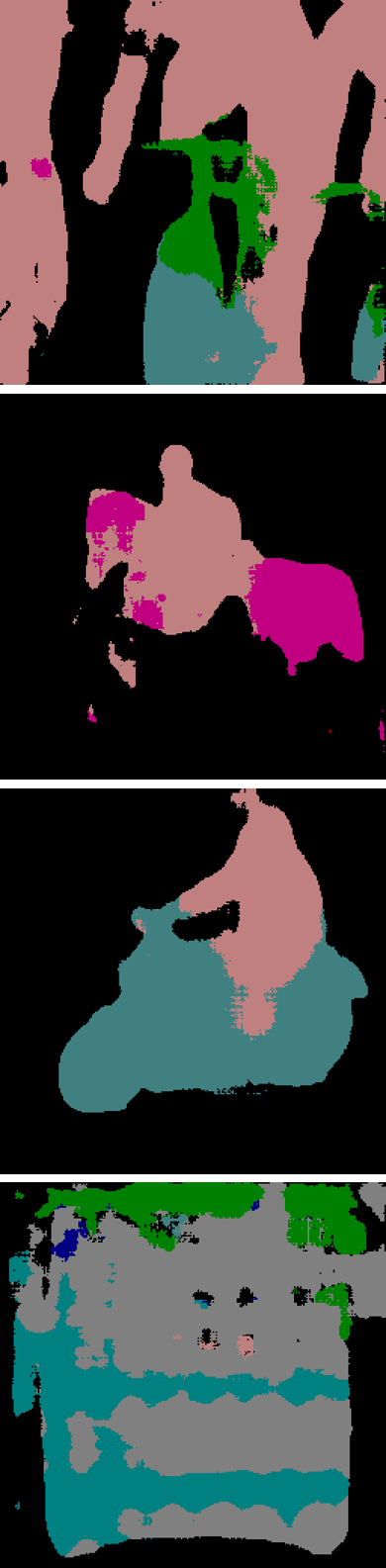}& 
 \includegraphics{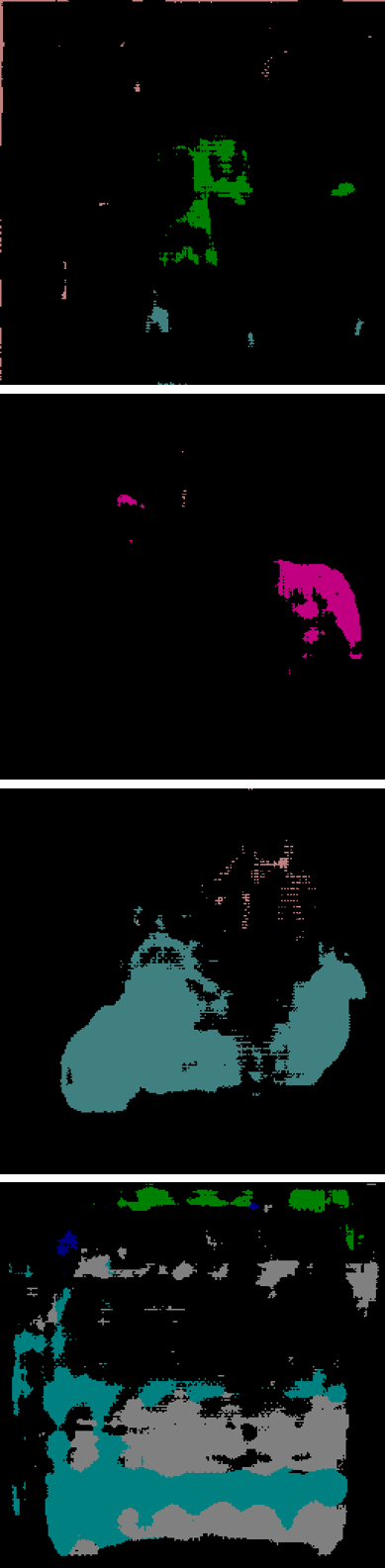}&
 \includegraphics{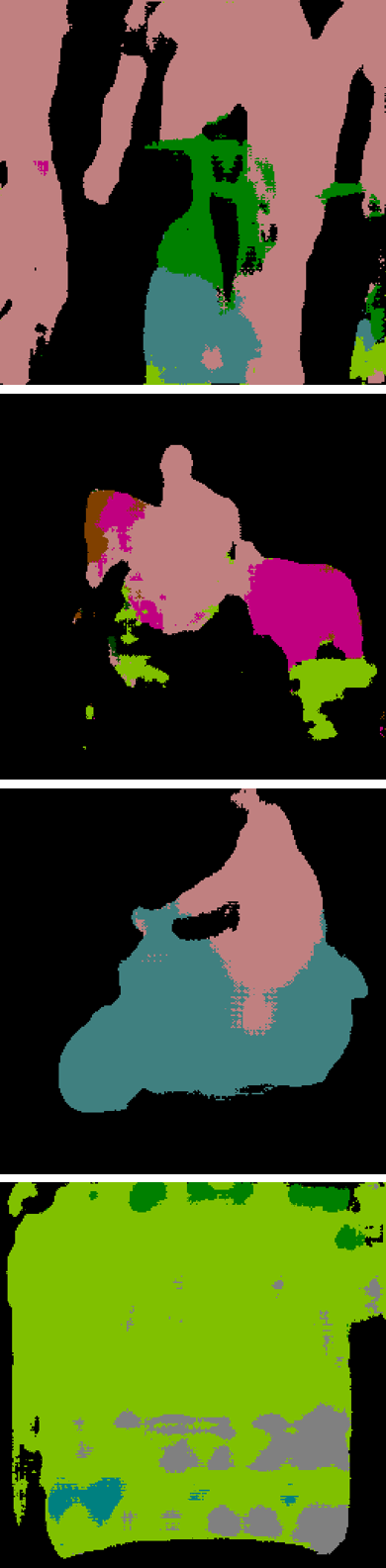}  
\end{tabular}%
}
\centering
\resizebox{0.7\textwidth}{!}{%
\begin{tabular}{ccccccccc}
 \cellcolor{bicycle} \textbf{bicycle} & \cellcolor{bus} \textbf{bus} & \cellcolor{car}\textbf{car} &  \cellcolor{horse} \textbf{horse} &  \cellcolor{motorbike} \textbf{motorbike}&  \cellcolor{person} \textbf{person} & \cellcolor{sheep} \textbf{sheep}  &\cellcolor{train} \textbf{train}  \\
\end{tabular}%
}

\caption{Visualizations of the segmentation maps for LwF, MAS and the resulting Frankenstein Networks of MAS $f_{1,0}^{17}$ and $f_{1,0}^{18}$. The predictions of $f_{1,0}^{17}$ and $f_{1,0}^{18}$ show that up until layer 17 the information for previously learned classes \textit{person} and \textit{horse} is still available, but is assigned to the \textit{background} in layer 18.}
\label{fig:gradcam}
\end{figure}

We use the Dr. Frankenstein tool set to measure the activation drift for each layer between the model before and after learning $T_{1}$, to investigate which layers are affected the most by the internal activation drift.
We follow the setup from \cref{fig:layer_stitch} without an additional stitching layer, meaning that the activations of the layer $n$ under examination $f_1$ are directly propagated to the layer $n+1$ in $f_0$. The resulting Frankenstein Network $f_{1,0}^n$ is then evaluated on the data of the first task (classes 0--15).
The mIoU relative to the initial performance on the first task is shown in \cref{fig:layer_stitch_results}. Overall, we observe that the activations of the early layers of the network (layers 0--4) stay very stable for every approach and that later layers are disproportionately affected by activation drift, especially the layers of decoder. A similar observation was made in prior work on image classification \cite{davari2021probing,NEURIPS2020_0607f4c7}.
This confirms that forgetting in CiSS is also mostly affecting deeper layers of the network. Furthermore, the analysis shows that EWC and MAS effectively prevent severe activation drift in the deeper layers of the encoder, dropping only to about 90\% of the initial mIoU on the disjoint task, compared to the 30\% of the stitched fine-tuning model.
This suggests that forgetting for EWC and MAS is less severe as accuracy in \cref{tab:overall_results} would reveal.
The reason for this could be two-fold: firstly the bad accuracy could be attributed to the classifier being biased towards new classes (task-recency bias) or secondly that the regularization methods fail to learn meaningful features that help to discriminate between old and new classes as they are never trained jointly. While a biased classifier is fixed more easily, inter-task confusion is a fundamental shortcoming of prior regularization methods \cite{lesort2020regularization}.\\ 
Another striking phenomenon is the severe change of activations at the third decoder layer (layer 18) that Fine-Tuning, MAS and EWC show on the \textit{Disjoint} task. The predictions of the specific Frankenstein Networks $f_{1,0}^{17}$ and $f_{1,0}^{18}$ in \cref{fig:gradcam}, show that $f_{1,0}^{17}$ is able to correctly classify old classes (\textit{bike}, \textit{person}) as such, but $f_{1,0}^{18}$ assigns the background class to these regions. Therefore, it can be concluded that the sudden activation change of MAS originates from the fact that features that were evidence for old classes in $f_0$ are now attributed evidence for the background class. This validates that the the semantic shift of the background class is mostly affecting the later layers of the decoder and that the features for old classes are in fact not forgotten, but assigned to the background class.
Similar observations can be made for EWC and Fine-Tuning. In \cref{CKAvsFrankenstein} measure the similarity with Centered Kernel Alignment to support our observations.
When completely avoiding the semantic background shift in the \textit{Full Disjoint} task, we observe that activation drift for the fine-tuned model is much more pronounced in the middle layers of the encoder (layer 8-15), which implies that the re-appearance of old classes, even though they are labelled as background, is mitigating the activation drift in the earlier layers of the model. 

\begin{figure}[]
\centering
\includegraphics[width=.5\textwidth]{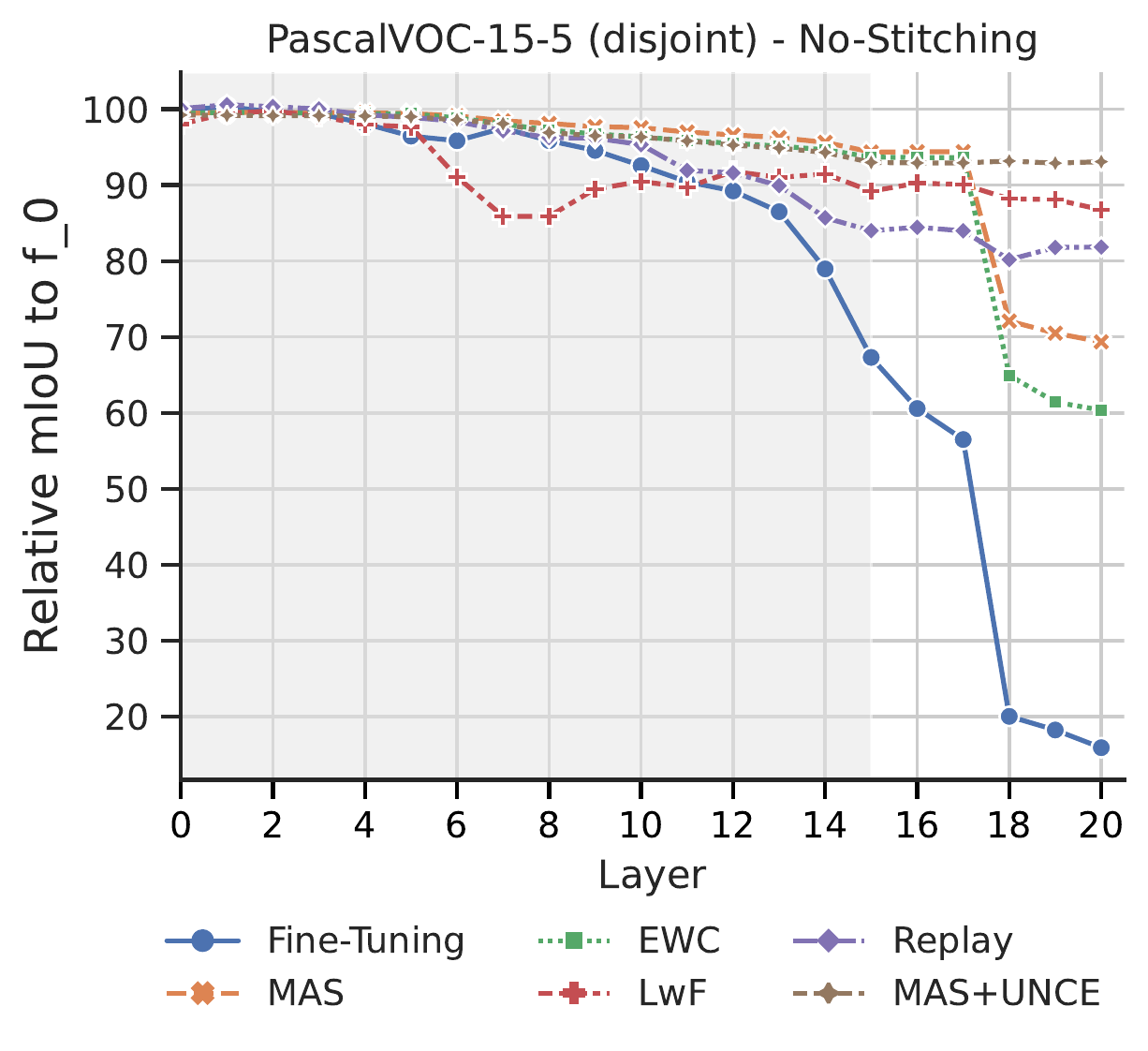}\hfill
\includegraphics[width=.5\textwidth]{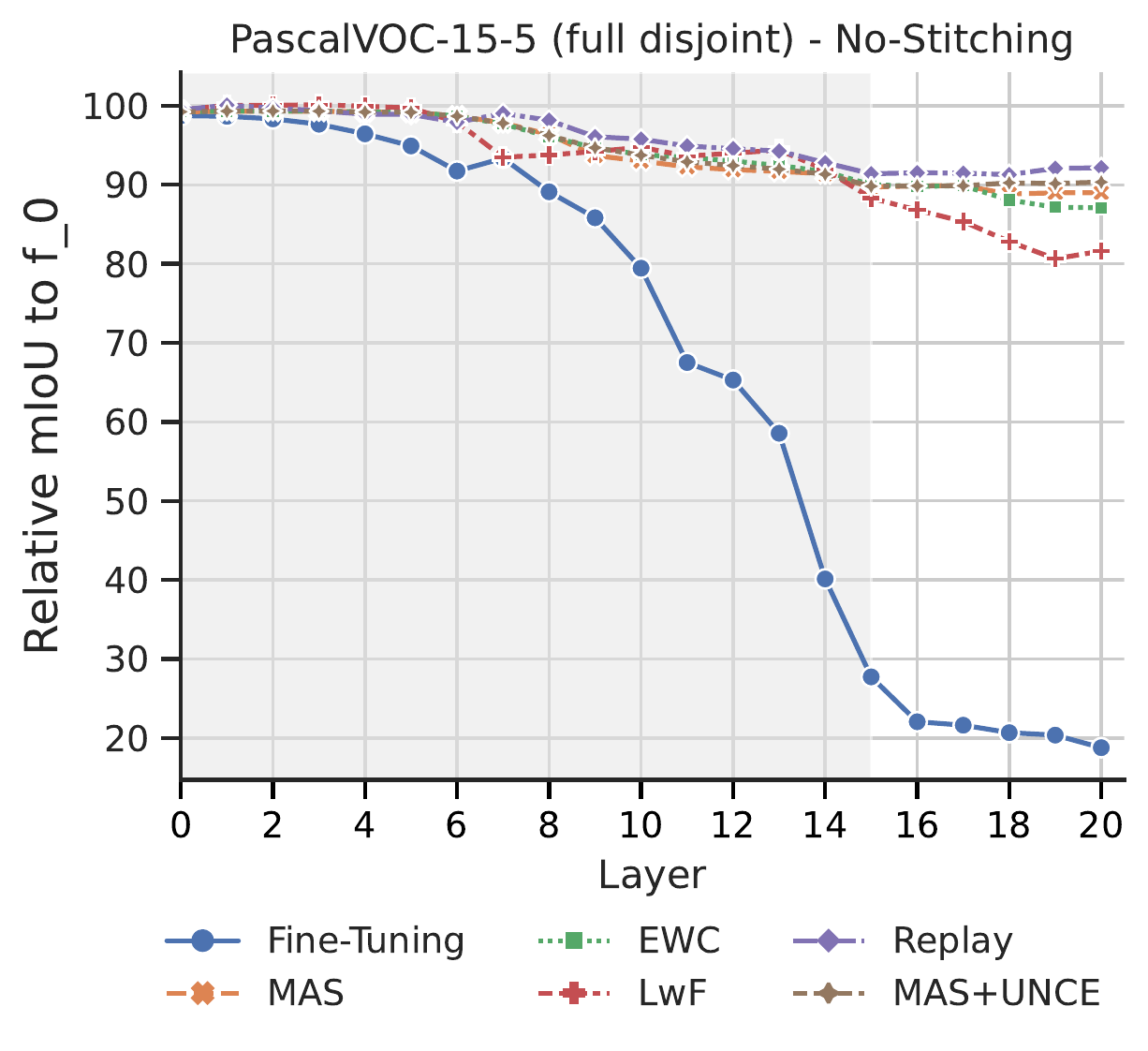}
\caption{Activation drift between $f_1$ to $f_0$ measured by relative mIoU on the first task of the Frankenstein Networks stitched together at specific layers (horizontal axis). The layers of the encoder are layer 0--15 (grey area), the decoder layers are 17-20 (white area). The activations in the early layers of the encoder stay very stable for all methods, whereas EWC, MAS and Fine-Tuning have a severe drift in activations in the decoder layers of the network, which is clear evidence that forgetting is mostly affecting later layers in the \textit{Disjoint} setting. }
\label{fig:layer_stitch_results}
\end{figure}

\subsection{The Impact of Inter-task Confusion on the Encoder}
\begin{figure}[]
    \centering
    \begin{subfigure}[t]{0.25\textwidth}
        \includegraphics[width=\textwidth]{img/confmat/MAS_confmat.pdf}
        \caption{MAS}
        \label{fig:conf_MAS2}
    \end{subfigure}
    \begin{subfigure}[t]{0.25\textwidth}
        \includegraphics[width=\textwidth]{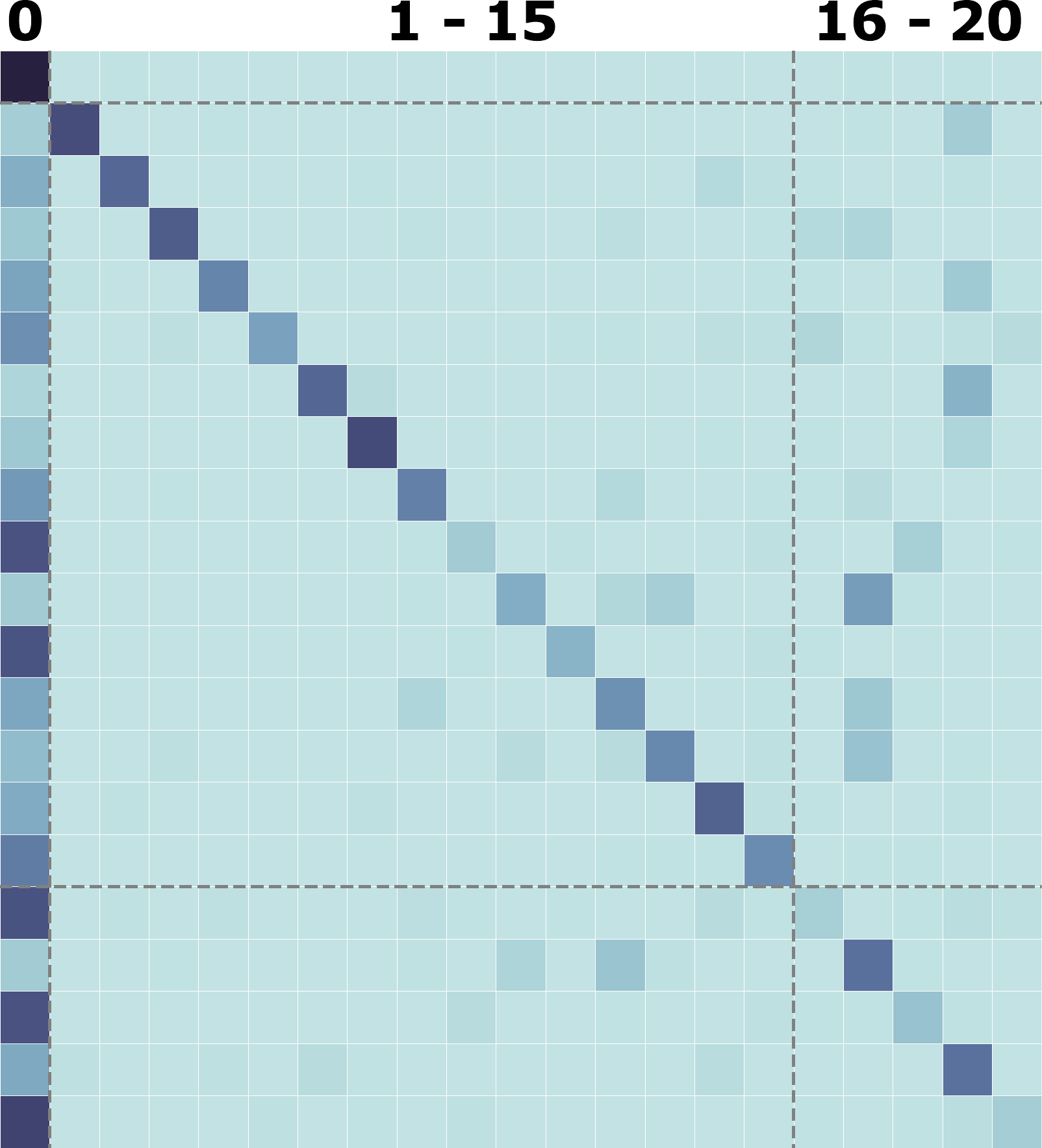}
        \caption{MAS-Retrain}
        \label{fig:conf_MAS_re-train}
    \end{subfigure}
    \begin{subfigure}[t]{0.25\textwidth}
        \includegraphics[width=\textwidth]{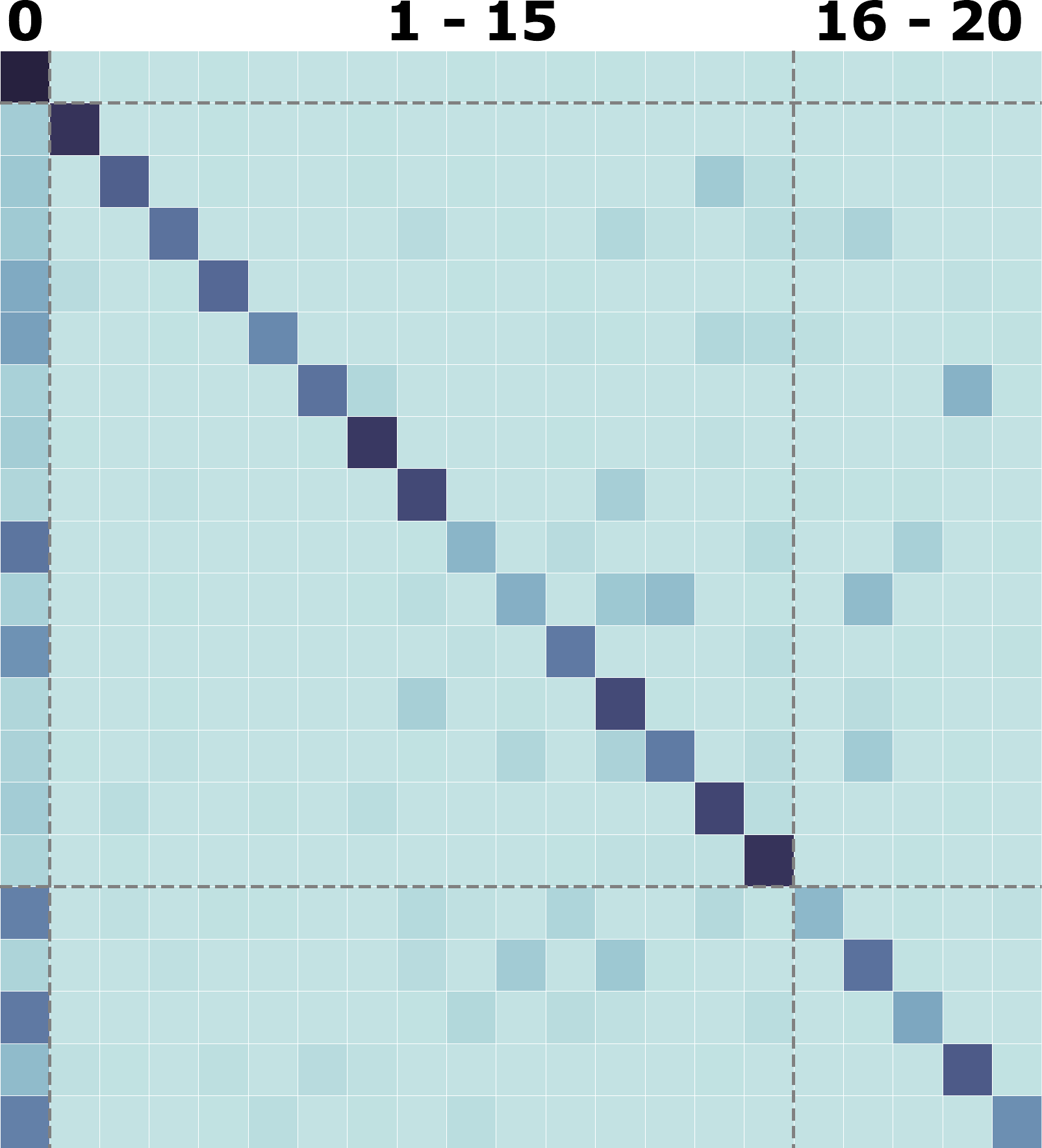}
        \caption{LwF-Retrain}
        \label{fig:conf_lwf_re-train}
    \end{subfigure}
    \caption{Confusion matrices before (a) and after b), c) retraining the decoder on all classes of PascalVoc2012.} \label{fig:ret-train_confmat}
\end{figure}

As we observed in \cref{activation_drift} that the early layers of a model trained with a continual learning method do not suffer from severe activation drift, in this experiment we investigate how useful the learned features of the encoder of the different methods are to discriminate between all classes. Therefore, we measure Decoder Retrain Accuracy, introduced in \cref{decoder_retrain}, for which the decoder of the model is retrained on all classes and subsequently evaluated on the test set.
The first observation to be made when looking at the retraining accuracy in \cref{tab:re-train}, is that all methods improve after decoder retraining, though EWC, MAS and Fine-Tuning show bigger improvements than LwF and Replay. This again confirms that forgetting in the encoder is not as severe for Fine-Tuning, EWC and MAS as the accuracy indicates. Furthermore, it also verifies that MAS and EWC are effectively preserving important features for old classes in the encoder, but that the biased decoder layer might wrongly attribute important features for old classes to the background class or new classes, which leads to a severe amount of misclassifications.\\
Still, as EWC, MAS and Fine-Tuning do not achieve a comparable mIoU as LwF or Replay after decoder retraining, it can be concluded that the learned features of the encoder are less useful for discriminating between all classes. Specifically the aforementioned related classes \textit{bus} (6), \textit{car} (7), \textit{boat} (5), \textit{train} (19), as well as \textit{cow} (10), \textit{horse} (13), \textit{sheep} (17) cannot be effectively classified after retraining, compare \cref{fig:ret-train_confmat}.
We hypothesize that Replay does not suffer from inter-task confusion as during optimization for the new classes, old classes are taken into account, so more discriminative features can be learned.
The same holds for LwF in the \textit{Overlapped} and \textit{Disjoint} setting, in which old classes are effectively replayed, by using soft-labels for old classes that are discovered in the background.

\subsection{Mitigating background bias and the task recency bias}
A simple method to reduce the recency bias in the classification layer that is used in class-incremental classification is to calculate the cross-entropy loss (CE) only for classes of the current training set \cite{masana2020class}. This enforces that errors are only back-propagated for probabilities that are related to the current set of classes:
\begin{equation}\label{eq:ce}
 \ell_{\text{ce}}(y, q) = -\frac{1}{I} \sum_{ i\in \mathcal{I} }{}\sum_{c \in \mathcal{C}_{t-1}}{}{y_{i, c}\log\left(  q_{i, c} \right)} 
\end{equation}
However, in the case of CiSS, this addition has proven to be less effective than the standard cross-entropy loss \cite{Douillard2020}. Therefore, an unbiased cross-entropy loss (UNCE) is proposed in \cite{Cermelli2020}, which accounts for the uncertainty of the content of the background class. This is achieved by comparing the pixels that are labelled as background with the probability of having either an old class or the background predicted by the model:
\begin{equation}\label{eq:unce_1}
 \ell_{\text{ce}}(y, q) = -\frac{1}{I} \sum_{ i\in \mathcal{I} }{}\sum_{c \in \mathcal{C}}{}{y_{i, c}\log\left(  \hat{q}_{i, c} \right)}   
\end{equation}
\begin{equation}\label{eq:unce_2}
 \hat{q}_{i, c} = \begin{cases}
      \sum_{k \in \mathcal{C}_{t-1}} q_{i, k} & \text{if $c = b$}\\
      q_{i, c} & \text{otherwise}\\
    \end{cases} 
\end{equation} 
In addition, Weight Normalization Layers \cite{NIPS2016_ed265bc9} were also successfully used in classification tasks to address the recency bias \cite{Lesort2021}. In the next experiment we study the impact of UNCE and UNCE combined with Weight Normalization to combat the recency and background bias in CiSS.\\
The results in \cref{tab:disjoint_UNCE} show that UNCE improves the accuracy for all approaches on the \textit{Disjoint} setting. Specifically, the prior regularization methods MAS and EWC show a significantly higher accuracy compared to the basic cross-entropy loss. This can be attributed to the fact that UNCE effectively mitigates the background bias, as we can see in the confusion matrix in \cref{fig:conf_MASUNCE} and the segmentation maps in \cref{fig:gradcam}. In addition, the severe activation drift that we observed in \cref{activation_drift} between layer 17 and 18 for MAS completely vanishes with the use of UNCE. Therefore, UNCE effectively resolves the confusion of the old classes with the background class.
This confirms the assumption that a major cause of forgetting in \cref{general_results} was in fact a bias of the classifier towards the background and the new classes.
However, the confusion matrix shows that while the background bias is severely reduced by using UNCE, the semantic confusion of old and new classes is amplified.\\  
In the \textit{Full Disjoint} setting the use of UNCE does not improve the performance as much as it does in the \textit{Disjoint} setting.
The reason is that in the \textit{Full Disjoint} setting the pixels of old classes do not re-occur and thus the de-biasing effect of UNCE is decreased. Therefore, the content of the background class plays an important role to mitigate forgetting.
Of the selected approaches, only Replay benefits from the addition of the Weight Normalization layer. Finally, we note that EWC and MAS, with the addition of UNCE, show competitive performance to the remaining approaches and more recent approaches like MiB \cite{Cermelli2020}, even without the use of knowledge-distillation or replay. However, we hypothesize that for longer task sequence and more classes MiB will outperform prior regularization methods, as they will not be able to learn discriminative features.

\begin{table}[]
\centering

\caption{Results on Pascal-15-5 in mIoU (\%) on the disjoint and full-disjoint settings with: Cross-Entropy Loss (CE), Unbiased Cross-Entropy (UNCE) and UNCE combined with a Weight Normalization (UNCE+WN). UNCE effectively reduces forgetting for all approaches, especially for EWC and MAS.}
\label{tab:disjoint_UNCE}
\resizebox{\textwidth}{!}{%
\begin{tabular}{l|ccccccccc|ccccccccc}
\multicolumn{1}{c|}{\textbf{}} & \multicolumn{9}{c|}{\textbf{PascalVoc 15-5 (disjoint)}} & \multicolumn{9}{c}{\textbf{PascalVoc 15-5 (full disjoint)}} \\ \hline
 & \multicolumn{3}{c|}{\textbf{CE}} & \multicolumn{3}{c|}{\textbf{UNCE}} & \multicolumn{3}{c|}{\textbf{UNCE+WN}} & \multicolumn{3}{c|}{\textbf{CE}} & \multicolumn{3}{c|}{\textbf{UNCE}} & \multicolumn{3}{c}{\textbf{UNCE+WN}} \\
\textbf{Method} & \textit{0-15} & \multicolumn{1}{c|}{\textit{16-20}} & \multicolumn{1}{c|}{\textit{all}} & \textit{0-15} & \multicolumn{1}{c|}{\textit{16-20}} & \multicolumn{1}{c|}{\textit{all}} & \textit{0-15} & \multicolumn{1}{c|}{\textit{16-20}} & \textit{all} & \textit{0-15} & \multicolumn{1}{c|}{\textit{16-20}} & \multicolumn{1}{c|}{\textit{all}} & \textit{0-15} & \multicolumn{1}{c|}{\textit{16-20}} & \multicolumn{1}{c|}{\textit{all}} & \textit{0-15} & \multicolumn{1}{c|}{\textit{16-20}} & \textit{all} \\ \hline

\multicolumn{1}{l|}{Fine-Tuning} & 4.6 & \multicolumn{1}{c|}{23.0} & \multicolumn{1}{c|}{9.0} & 10.4 & \multicolumn{1}{c|}{21.8} & \multicolumn{1}{c|}{13.1} & 16.5 & \multicolumn{1}{c|}{21.6} & 17.7 & 5.1 & \multicolumn{1}{c|}{16.3} & \multicolumn{1}{c|}{7.8} & 6.0 & \multicolumn{1}{c|}{15.5} & \multicolumn{1}{c|}{8.3} & 7.7 & \multicolumn{1}{c|}{15.2} & 9.5 \\ 
\multicolumn{1}{l|}{EWC \cite{Kirkpatrick2015}} & 28.1 & \multicolumn{1}{c|}{10.1} & \multicolumn{1}{c|}{23.8} & 48.2 & \multicolumn{1}{c|}{11.6} & \multicolumn{1}{c|}{39.4} & 17.0 & \multicolumn{1}{c|}{9.5} & 15.2 & 35.2 & \multicolumn{1}{c|}{10.9} & \multicolumn{1}{c|}{29.4} & 41.1 & \multicolumn{1}{c|}{9.8} & \multicolumn{1}{c|}{33.6} & 34.8 & \multicolumn{1}{c|}{9.7} & 28.8 \\
\multicolumn{1}{l|}{MAS \cite{Aljundi2018_MAS}} & 30.6 & \multicolumn{1}{c|}{12.9} & \multicolumn{1}{c|}{26.4} & 45.8 & \multicolumn{1}{c|}{14.4} & \multicolumn{1}{c|}{38.3} & 41.0 & \multicolumn{1}{c|}{13.9} & 34.6 & 35.6 & \multicolumn{1}{c|}{12.9} & \multicolumn{1}{c|}{30.2} & 39.1 & \multicolumn{1}{c|}{12.3} & \multicolumn{1}{c|}{32.7} & 32.5 & \multicolumn{1}{c|}{11.8} & 27.6 \\ 
\multicolumn{1}{l|}{LwF \cite{Li2018}} & 44.4 & \multicolumn{1}{c|}{25.4} & \multicolumn{1}{c|}{39.9} & 45.3 & \multicolumn{1}{c|}{22.9} & \multicolumn{1}{c|}{40.0} & 46.6 & \multicolumn{1}{c|}{19.7} & 40.2 & 35.9 & \multicolumn{1}{c|}{12.6} & \multicolumn{1}{c|}{30.4} & 38.0 & \multicolumn{1}{c|}{13.8} & \multicolumn{1}{c|}{32.2} & 38.8 & \multicolumn{1}{c|}{13.3} & 32.8 \\ 
\multicolumn{1}{l|}{Replay} & 42.2 & \multicolumn{1}{c|}{29.1} & \multicolumn{1}{c|}{39.1} & 47.2 & \multicolumn{1}{c|}{31.4} & \multicolumn{1}{c|}{43.5} & 48.1 & \multicolumn{1}{c|}{31.9} & 44.3 & 48.2 & \multicolumn{1}{c|}{28.8} & \multicolumn{1}{c|}{43.6} & 47.7 & \multicolumn{1}{c|}{28.0} & \multicolumn{1}{c|}{43.0} & 48.8 & \multicolumn{1}{c|}{28.5} & 44.0 \\ \hline
\multicolumn{1}{l|}{MiB \cite{Cermelli2020}} & - & \multicolumn{1}{c|}{-} & \multicolumn{1}{c|}{-} & 48.6 & \multicolumn{1}{c|}{21.7} & \multicolumn{1}{c|}{42.2} & 49.4 & \multicolumn{1}{c|}{24.1} & \multicolumn{1}{c|}{43.3} & - & \multicolumn{1}{c|}{-} & \multicolumn{1}{c|}{-} & 47.6 & \multicolumn{1}{c|}{19.7} & \multicolumn{1}{c|}{41.0} & 48.6 & \multicolumn{1}{c|}{20.7} & 42.0 \\
\end{tabular}%
}
\end{table}

\begin{minipage}[]{0.55\textwidth}
		\captionof{table}{Decoder Retraining Results on Pascal-VOC. $\text{mIoU}_I$ and $\text{mIoU}_R$ denote the mIoU (\%) before and after retraining.}
		\label{tab:re-train}
		\centering
		\resizebox{\textwidth}{!}{%
			\begin{tabular}{lcccccc}
				\multicolumn{7}{c}{\textbf{PascalVoc 15-5 - Decoder Retraining Accuracy}} \\ \hline
				\multicolumn{1}{l|}{} & \multicolumn{2}{c|}{\textbf{Overlapped}} & \multicolumn{2}{c|}{\textbf{Disjoint}} & \multicolumn{2}{c}{\textbf{Full Disjoint}} \\
				\multicolumn{1}{l|}{\textbf{Method}} & \textit{$\text{mIoU}_I$} & \multicolumn{1}{c|}{\textit{$\text{mIoU}_R$}} & \textit{$\text{mIoU}_I$} & \multicolumn{1}{c|}{\textit{$\text{mIoU}_R$}} & \textit{$\text{mIoU}_I$} & \textit{$\text{mIoU}_R$} \\ \hline
				
				\multicolumn{1}{l|}{Fine-Tuning} & 8.8 & \multicolumn{1}{c|}{28.0} & 9.0 & \multicolumn{1}{c|}{27.9} & 7.8 & 22.0  \\ 
				\multicolumn{1}{l|}{MAS \cite{Aljundi2018_MAS}} & 21.0 & \multicolumn{1}{c|}{34.3} & 26.4 & \multicolumn{1}{c|}{36.2} & 30.2 & 37.3  \\ 
				\multicolumn{1}{l|}{EWC \cite{Kirkpatrick2015}} & 21.0 & \multicolumn{1}{c|}{34.3} & 23.8 & \multicolumn{1}{c|}{35.1} & 29.4 & 36.9  \\ 
				\multicolumn{1}{l|}{LwF \cite{Li2018}} & 41.6 & \multicolumn{1}{c|}{45.3} & 39.9 & \multicolumn{1}{c|}{43.3} & 30.4 & 38.1  \\ 
				\multicolumn{1}{l|}{Replay} & 39.0 & \multicolumn{1}{c|}{42.6} & 39.1 & \multicolumn{1}{c|}{42.9} & 43.6 & 45.6  \\ \hline
				\multicolumn{1}{l|}{Offline} & 53.8 & \multicolumn{1}{c|}{54.6} & 53.8 & \multicolumn{1}{c|}{54.6} & 53.8 & 54.6  \\

			\end{tabular}%
	}

\end{minipage}
\hfill
\begin{minipage}[]{0.35\textwidth}
		\captionof{table}{Classification Results on PascalVoc-15-5.}
		\label{tab:classifciation}
		\centering
		\resizebox{\textwidth}{!}{%
			\begin{tabular}{lccc}
				\multicolumn{4}{c}{\textbf{PascalVoc 15-5 Classification}} \\ \hline
				\multicolumn{1}{l|}{} & \multicolumn{3}{c}{\textbf{Full Disjoint}} \\
				\multicolumn{1}{l|}{\textbf{Method}} & \textit{0-15} & \multicolumn{1}{c|}{\textit{16-20}} & \textit{all} \\ \hline
				\multicolumn{1}{l|}{Fine-Tuning} & 13.6 & \multicolumn{1}{c|}{27.6} & 17.1  \\ 
				\multicolumn{1}{l|}{MAS \cite{Aljundi2018_MAS}} & 32.0 & \multicolumn{1}{c|}{27.8} & 31.0  \\ 
				\multicolumn{1}{l|}{EWC \cite{Kirkpatrick2015}} & 27.2 & \multicolumn{1}{c|}{25.4} & 26.8  \\
				\multicolumn{1}{l|}{LwF \cite{Li2018}} & 39.6 & \multicolumn{1}{c|}{32.7} & 37.9  \\ 
				\multicolumn{1}{l|}{Replay} & 42.1 & \multicolumn{1}{c|}{34.2} & 40.1  \\ 
				\hline
				\multicolumn{1}{l|}{Offline} & 51.3 & \multicolumn{1}{c|}{54.8} & 52.2
			\end{tabular}%
		}
\end{minipage}

\subsection{The Role of the Background Class to overcome forgetting}
The prior observations show that in CiSS the semantic shift of the background class is a major cause of a rapid drop in performance if not addressed correctly. However, if the uncertainty of the content of the background class is taken into account by either UNCE, Knowledge Distillation or both, the appearance of old classes in the background can to some extent be used for replay. In the experiments of the \textit{Full Disjoint} setting we see that once classes do not reoccur, these methods are less effective, whereas explicit replay of classes benefits from avoiding the semantic shift. 
The ranking of the methods of \textit{Full Disjoint} setting in CiSS is also similar to the ranking of the same methods for class-incremental image classification, indicating that the discrepancy in performance of LwF and replay in image classification is due to the missing background class. Looking at it the other way around, this could also mean that introducing an out-of-set class for image classification could help to reduce forgetting in the class-incremental setting without requiring explicit replay via stored samples, as the re-appearing classes in the out-of-set-class play a similar role as explicit replay. 

\section{Conclusion}
We studied the major causes of catastrophic forgetting in CiSS, answering how it manifests itself in the hidden representations of the network and how the background class both causes severe forgetting and decreases activation drift. Using representational similarity techniques, we demonstrated that forgetting is concentrated at higher layers and that re-appearing classes mitigate activation drift in the encoder even when they are labelled as background. Moreover, we show that EWC and MAS are effectively reducing representational drift in the later layers of the encoder, but suffer from severe background and recency bias, which leads to the sudden drop in accuracy. These biases manifest themselves in deeper layers of the networks by assigning previous discriminating features for the previous classes to the background class or visually related classes.\\
The background bias can be effectively alleviated using an unbiased cross-entropy loss, which leads to a significant improvement for all methods, when classes re-appear in the background of new training data. Finally, we find that only methods that in some form replay old classes during training of new classes can learn to correctly discriminate between all classes after incremental training, as otherwise the model fails to learn to discriminate between new and old classes that share similar visual features.
Overall, the results of our work provide a foundation for deeper understanding of the principles of forgetting in CiSS and open the door to future directions to explore methods for its mitigation. 

\section*{Acknowledgment}
The research leading to these results is funded by the German Federal Ministry for Economic Affairs and Climate Action within the project “KI Delta Learning“ (Förderkennzeichen 19A19013T). The authors would like to thank the consortium for the successful cooperation.

\clearpage
\bibliographystyle{splncs}
\bibliography{sources}

\clearpage

\appendix

\section{Reproducibility}
\subsection{Training Protocol}
For the experiments we follow the same train, test and validation splits as \cite{Cermelli2020}. Instead of using ImageNet pre-trained models, we use the same randomly initialized ERFNet in all our experiments. For optimization we utilize Stochhastic Gradient Descent in combination with wight decay factor of $3\times 10^{-4}$, momentum of 0.9 and an initial learning rate of 0.07 for the first task and 0.0005 for the second task. The learning rate is divided by 2 if the validation loss is not reducing for 8 consecutive epochs. We train the model in each task for 100 epochs with a batch size of 16.
After training on the entire training sequence the model is evaluated on the validation set of PascalVoc2012.
During training we use the same augmentations as \cite{Cermelli2020}, but use the implementations of Albumentations.

\subsection{Continual Hyperparameter Selection}
To select the hyperparameters for the continual learning methods we follow the Continual Hyperparameter Framework of \cite{aljundi_survey}, by firstly choosing the appropriate learning rate that achieves the highest accuracy on the new task and consecutively tuning the specific continual learning hyperparameters. This lead to the following hyperparameters for the respective methods:
\begin{itemize}
    \item Fine-Tuning: $\text{lr}_0 = 0.07$,\quad $\text{lr}_1 = 5\times 10^{-4}$
    \item EWC\footnote{In order to use such a high value for $\lambda$ we clip the gradient norm at a value of 10 to avoid exploding gradients.}: $\lambda = 10000$
    \item MAS: $\lambda = 5000$
    \item LwF: $\lambda = 6$,\quad $T=2$
    \item MiB: $\lambda = 25$,\quad $\alpha = 1$
\end{itemize}
For all the continual learning methods, we used the same $\text{lr}_0 = 0.07$ and $\text{lr}_1 = 5\times 10^{-4}$ as for Fine-Tuning. In the experiments with UNCE and Weight Normalization we use the exact same parameters as before.

\subsection{Implementation:}
All of our experiments are conducted using PyTorch in combination with Pytorch Lightning. We use the original PyTorch implementation of ERFNet provided by \cite{erfnet} which can be found at: \href{https://github.com/Eromera/erfnet}{https://github.com/Eromera/erfnet}. The implementation of the unbiased cross-entropy loss is taken from \cite{Cermelli2020}, which can be found at: \href{https://github.com/fcdl94/MiB}{https://github\allowbreak .com/fcdl94/MiB}. For the continual learning algorithms we adapt the implementation of \cite{masana2020class} for the use in semantic segmentation. The reference code can be found at: \href{https://github.com/mmasana/FACIL}{https://github.com/mmasana/FACIL}. 

\subsection{Decoder Retraining}
For Decoder Retraining we freeze all layers of the encoder of $f_1$ and set them in evaluation mode. We only train the decoder on top of the encoder again with SGD weight decay factor of $3\text{e-}4$, momentum of 0.9. We again use a initial learning rate of 0.07, but replace the learning rate schedule of the previous experiments with a cosine annealing learning rate schedule. The model is trained for 30 epochs on the PascalVoc-2012 training set.

\section{Results for different CNN-Architectures}
We further also validated those findings on the widely adopted UNet and DeepLabv3 each with ResNet101 Backbones. When using the cross-entropy loss (CE) the results are of the models is equal, because information of old classes is completely overwritten, as the models are optimize to assign the background class to old classes. However, when using UNCE we see that the models with higher learning capacity can retain much more of the information of old classes than the smaller ERFNet. When looking at the internal representation shifts, DeepLabv3 beahves similar to ERFNet, as it retains high similarity throughout the network and suffers from a big drop in similarity in the layers of the decoder. On the other hand, for UNet even early layers (from layer 10) are affected by a representation shift. A likely explanation are the skip-connections that connect early layers to later layers in the network.

\begin{figure}[]
	\centering
	\label{tab:overall_results}
	\resizebox{0.65\textwidth}{!}{%
		\begin{tabular}{lcccccccccccccccccc}
			\multicolumn{7}{c|}{\textbf{PascalVoc 15-5 - (disjoint)}} & \multicolumn{6}{c}{\textbf{PascalVoc 15-5 - (full-disjoint)}} \\ \hline
			\multicolumn{1}{l|}{} & \multicolumn{3}{c|}{\textbf{CE}} & \multicolumn{3}{c|}{\textbf{UNCE}} & \multicolumn{3}{c|}{\textbf{CE}} & \multicolumn{3}{c}{\textbf{UNCE}}\\
			
			\multicolumn{1}{l|}{\textbf{Method}} & \textit{0-15} & \multicolumn{1}{c|}{\textit{16-20}} &  \multicolumn{1}{c|}{\textit{all}} & \textit{0-15} & \multicolumn{1}{c|}{\textit{16-20}} & \multicolumn{1}{c|}{\textit{all}}  & \textit{0-15} & \multicolumn{1}{c|}{\textit{16-20}} &  \multicolumn{1}{c|}{\textit{all}} & \textit{0-15} & \multicolumn{1}{c|}{\textit{16-20}} & \textit{all} \\ \hline
			\multicolumn{1}{l|}{ERFNet} & 4.6 & \multicolumn{1}{c|}{23.0} & \multicolumn{1}{c|}{9.0}  & 10.4& \multicolumn{1}{c|}{21.8} & \multicolumn{1}{c|}{13.1} & 5.1 & \multicolumn{1}{c|}{16.3} & \multicolumn{1}{c|}{7.8} &  6.0& \multicolumn{1}{c|}{15.5} & 8.3 \\
			\multicolumn{1}{l|}{DeepLabV3} & 4.6 & \multicolumn{1}{c|}{24.7} & \multicolumn{1}{c|}{9.4} &  21.4& \multicolumn{1}{c|}{23.7} & \multicolumn{1}{c|}{21.9} & 5.7 & \multicolumn{1}{c|}{16.8} & \multicolumn{1}{c|}{7.9} &  5.7& \multicolumn{1}{c|}{16.2} & 8.2 \\
			\multicolumn{1}{l|}{UNet} & 4.6 & \multicolumn{1}{c|}{25.6} & \multicolumn{1}{c|}{9.6} &  12.7& \multicolumn{1}{c|}{25.3} & \multicolumn{1}{c|}{15.7} & 5.3 & \multicolumn{1}{c|}{16.2} & \multicolumn{1}{c|}{7.9} & 6.1  & \multicolumn{1}{c|}{16.6} & 8.6 \\
	\end{tabular}}
\end{figure}

\begin{figure}[]
	\centering
	\includegraphics[width=0.9\textwidth]{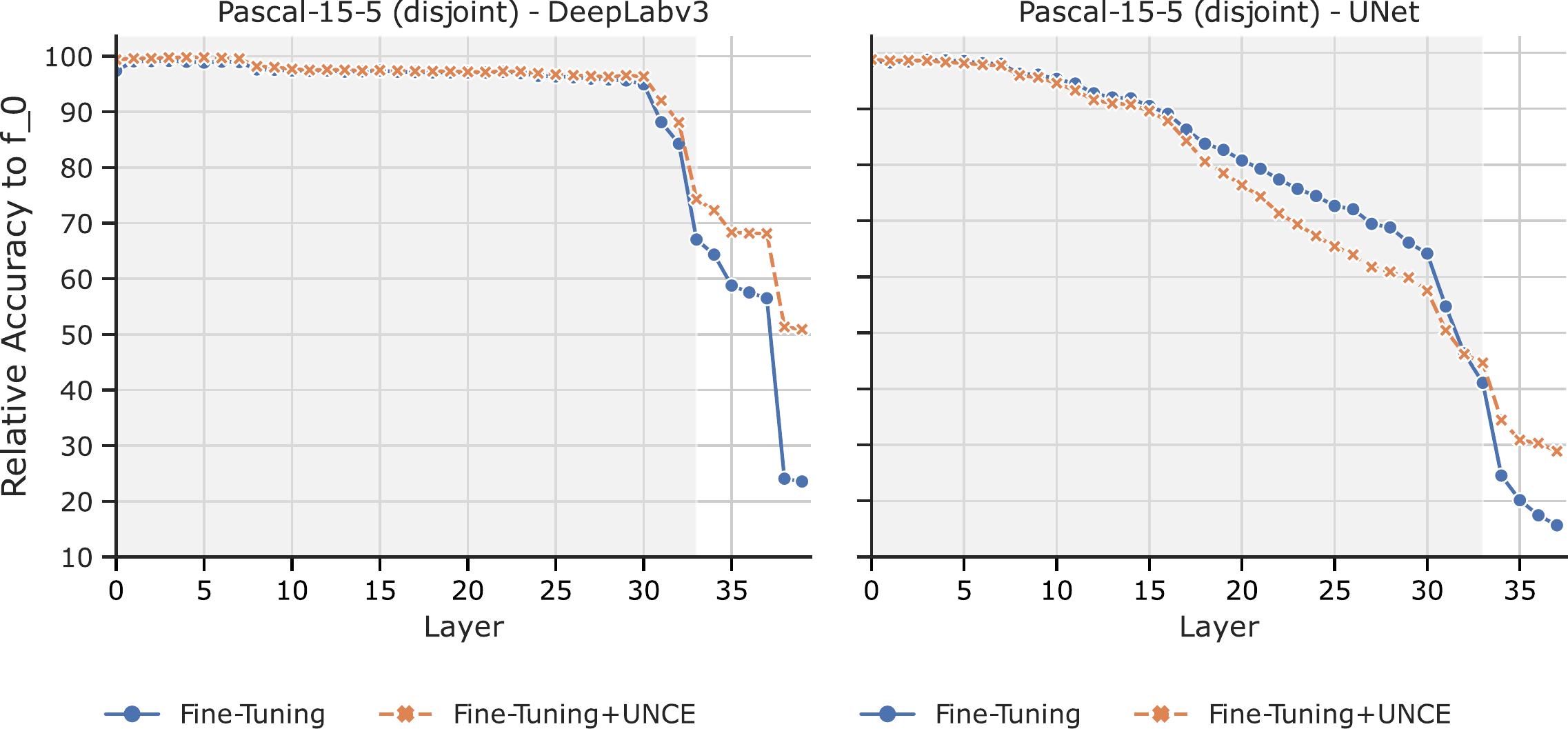}
	\label{fig:layer_stitch_results}
\end{figure}

\section{Dr. Frankenstein vs. Centered Kernel Alignment}\label{CKAvsFrankenstein}
\begin{figure}[]
\centering
\includegraphics[width=.45\textwidth]{img/stiching/UNCE_disjoint_No-Stitching.pdf}\hfill
\includegraphics[width=.45\textwidth]{img/stiching/UNCE_full_disjoint_No-Stitching.pdf}
\includegraphics[width=.45\textwidth]{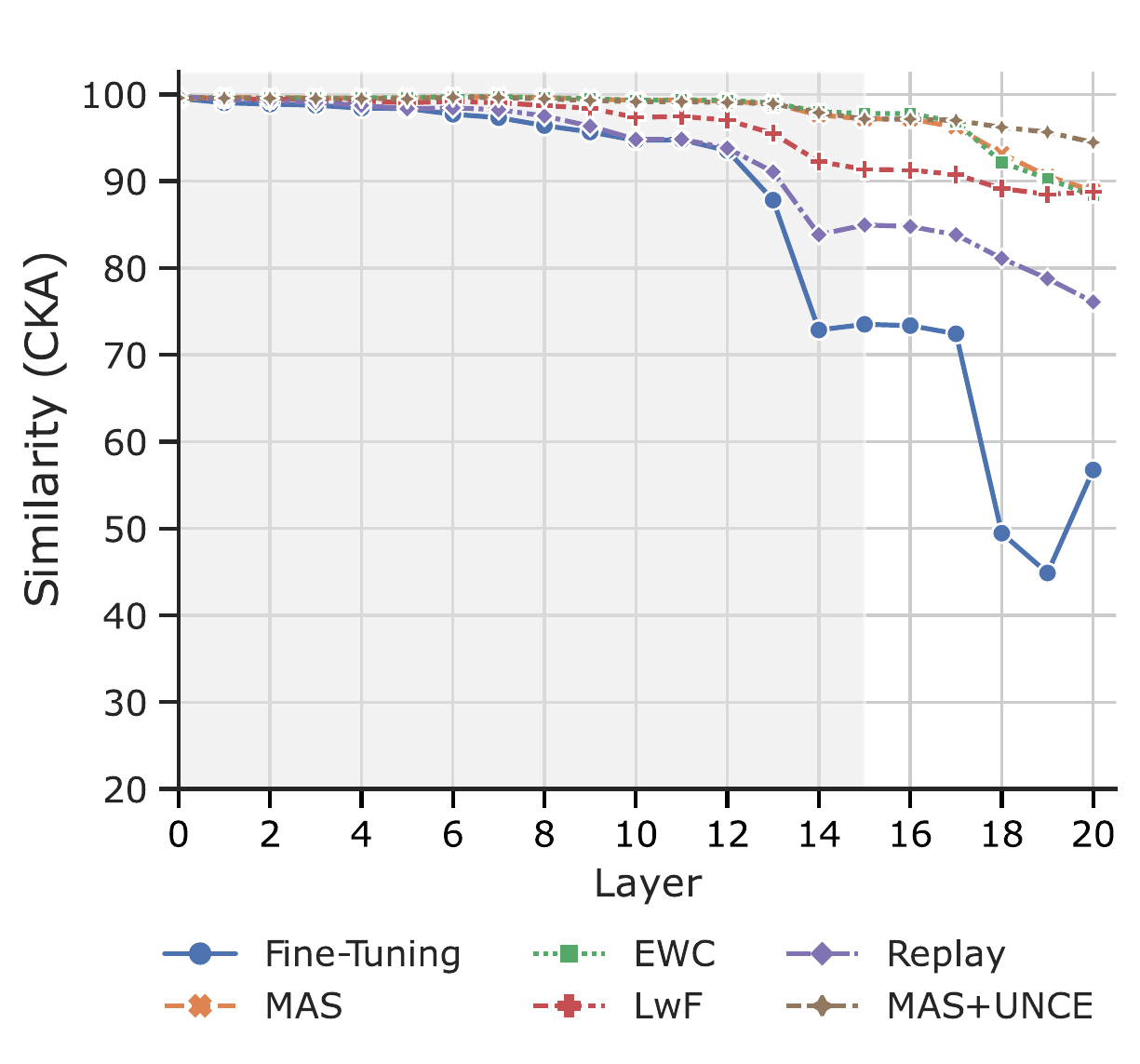}\hfill
\includegraphics[width=.45\textwidth]{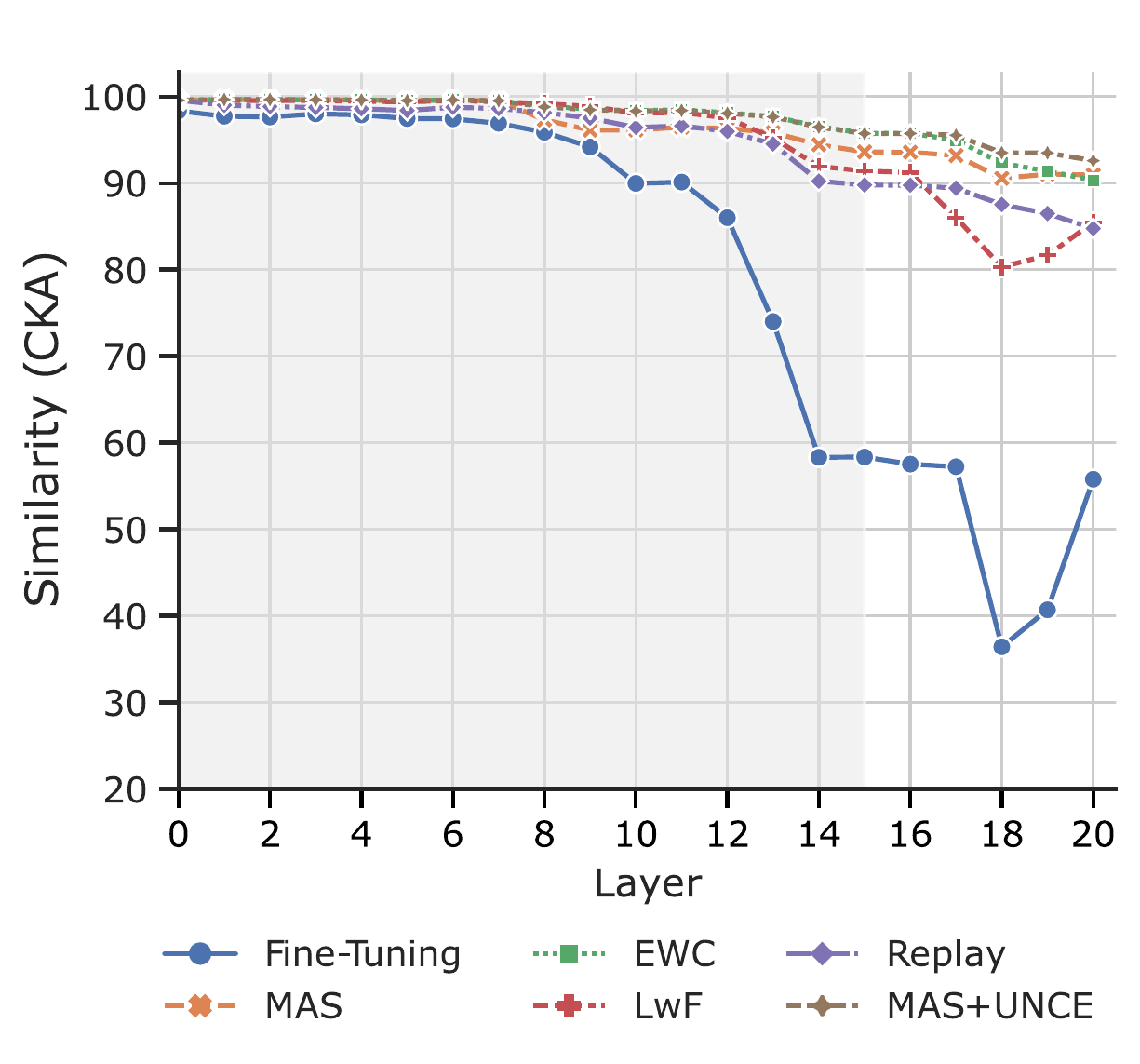}
\caption{Activation drift between $f_1$ to $f_0$ measured by relative mIoU on the first task of the Frankenstein Networks stitched together at specific layers (horizontal axis). The layers of the encoder are layer 0--15 (grey area), the decoder layers are 17-20 (white area). The activations in the early layers of the encoder stay very stable for all methods, whereas EWC, MAS and Fine-Tuning have a severe drift in activations in the decoder layers of the network, which is clear evidence that forgetting is mostly affecting later layers in the \textit{Disjoint} setting. }
\label{fig:cka_stitch}
\end{figure}

Centered Kernel Alignment (CKA) \cite{Kornblith2019} is a similarity index that measures the similarity between internal representation of neural networks. In continual learning CKA has recently been used to measure the activation drift during of the intermediate layers of a neural network cause of forgetting in CNNs \cite{davari2021probing,ramasesh2021anatomy}. Csisz\'{a}rik \etal \cite{Csiszarik2021} investigated the relationship between representational similarity that is measured by CKA and functional similarity measured by Dr. Frankenstein. They demonstrate that a high CKA score between activations does not infer that the networks have functional similarity. Meaning that high functional similarity can be retained while simultaneously the representational similarity measured by CKA decreases. In \cref{fig:cka_stitch} we compare the functional similarity measured by Dr. Frankenstein and the representational similarity measured by CKA. In line with previous assumptions the CKA analysis also confirm that the representations of the first 4 layers are not affected by representational drift. However, in the decoder we observe that the representational drift between $f_0$ and $f_1$ for EWC and MAS is not as pronounced as for the measured by Dr. Frankenstein. This indicates that the activations only undergo a minor change in terms of representational distance, but that these minor activation changes have a severe impact on the functional similarity of the model. 

\section{Visualization of the Bias Values of the Classification Layer}
\begin{figure}[]
  \centering
  \includegraphics[width=\textwidth]{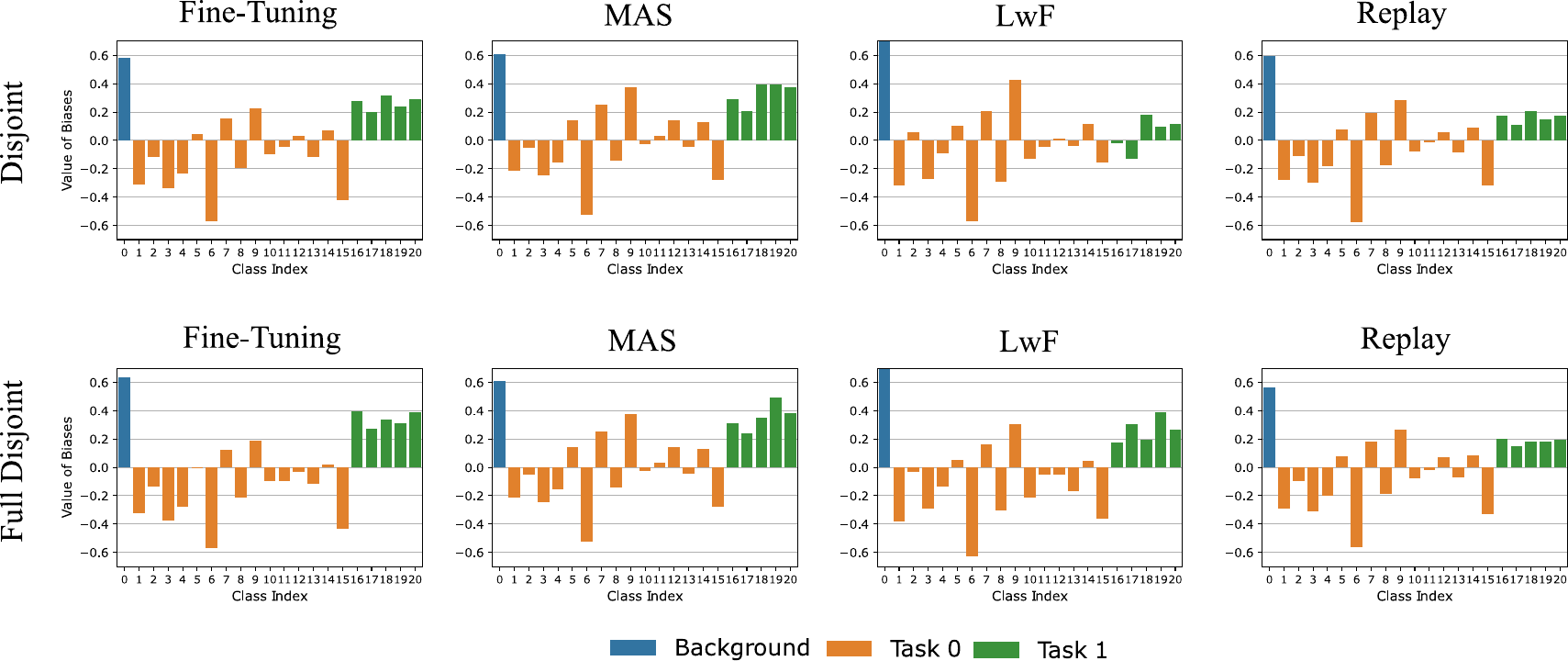}
  \caption{Bias Values of the final convolutional layer after training on the \textit{Disjoint} and \textit{Full Disjoint} PascalVoc-15-5 split.}
  \label{fig:bias_plots}
\end{figure}

The bias towards most recent classes and the background manifests itself in the bias values of the final convolution layer. The visualization in \cref{fig:bias_plots} demonstrates that all methods have obvious increased bias values for the background class and classes of $T_1$, with the exception of LwF in the \textit{Disjoint} setting. However, once the classes disappear from the background of the images the LwF shows a similar increased bias values.

\section{Confusion Matrices PascalVoc-15-5}
In the following we show all the confusion matrices for the \textit{Overlapped}, \textit{Disjoint} and \textit{Full-Disjoint} setting. The confusion matrices clearly illustrate that in the \textit{Full-Disjoint} the background bias is much less pronounced whereas the class confusion is more severe. 

\begin{figure}[]
\begin{tabular}{ccc}
\multicolumn{3}{c}{\textbf{Fine-Tuning}} \\
    \centering
    \begin{subfigure}[t]{0.28\textwidth}
        \includegraphics[width=\textwidth]{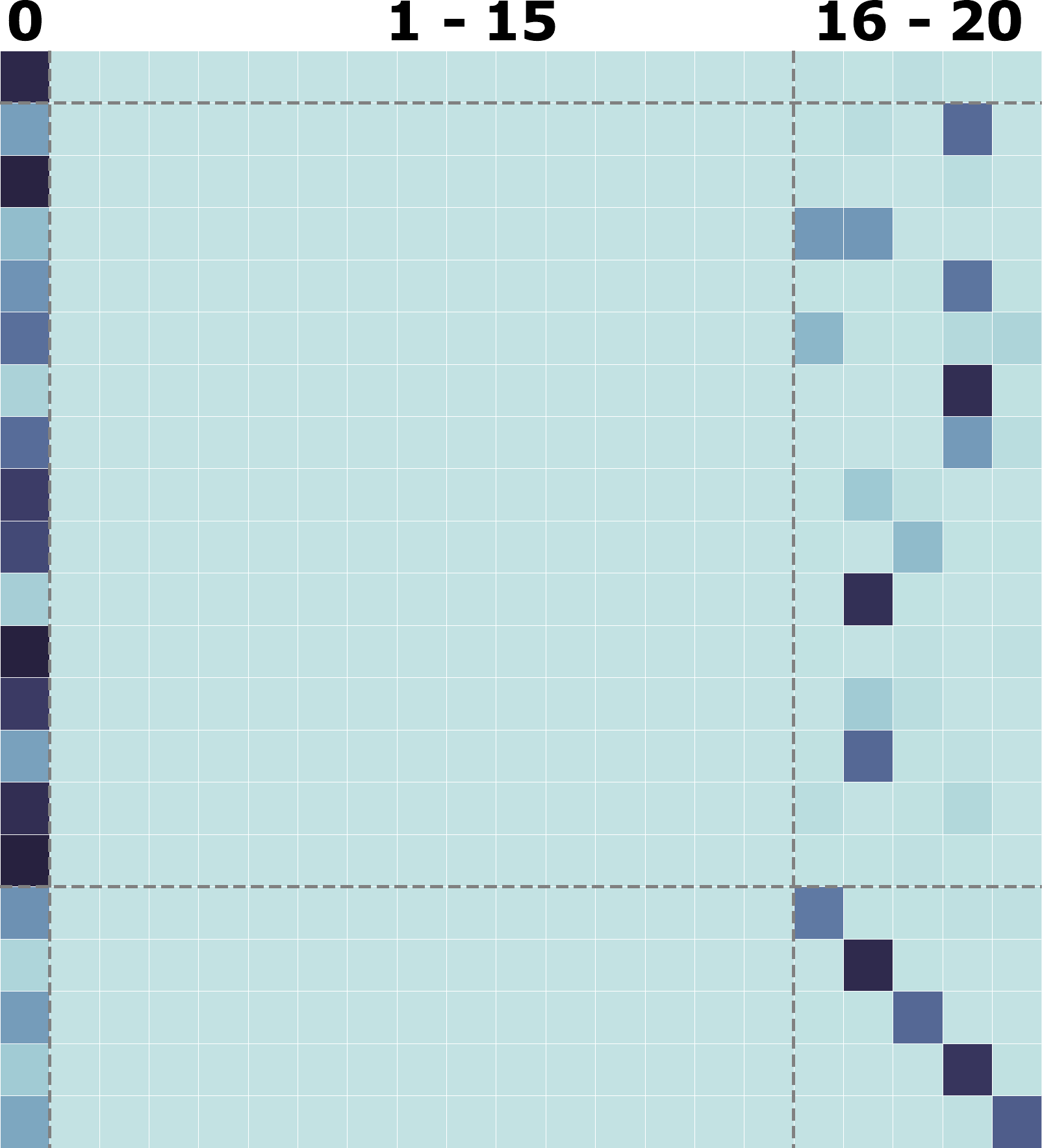}
        \caption{overlapped}
    \end{subfigure}    
    &  
        \begin{subfigure}[t]{0.28\textwidth}
        \includegraphics[width=\textwidth]{img/confmat/Fine-Tuning_confmat.pdf}
        \caption{disjoint}
    \end{subfigure}
    & 
    \begin{subfigure}[t]{0.28\textwidth}
        \includegraphics[width=\textwidth]{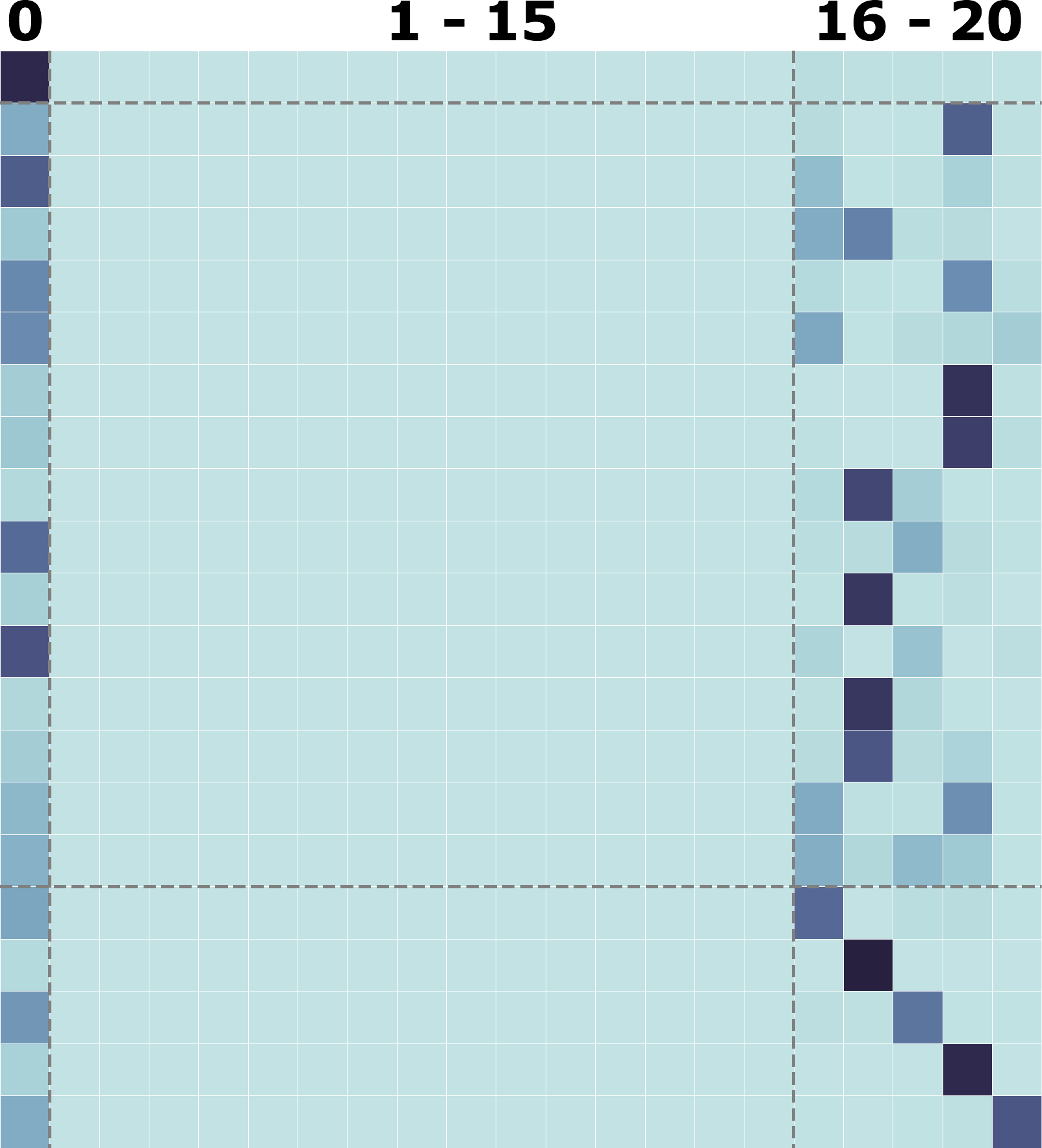}
        \caption{full-disjoint}
    \end{subfigure}
\end{tabular}%

\begin{tabular}{ccc}
\multicolumn{3}{c}{\textbf{MAS}} \\
    \centering
    \begin{subfigure}[t]{0.28\textwidth}
        \includegraphics[width=\textwidth]{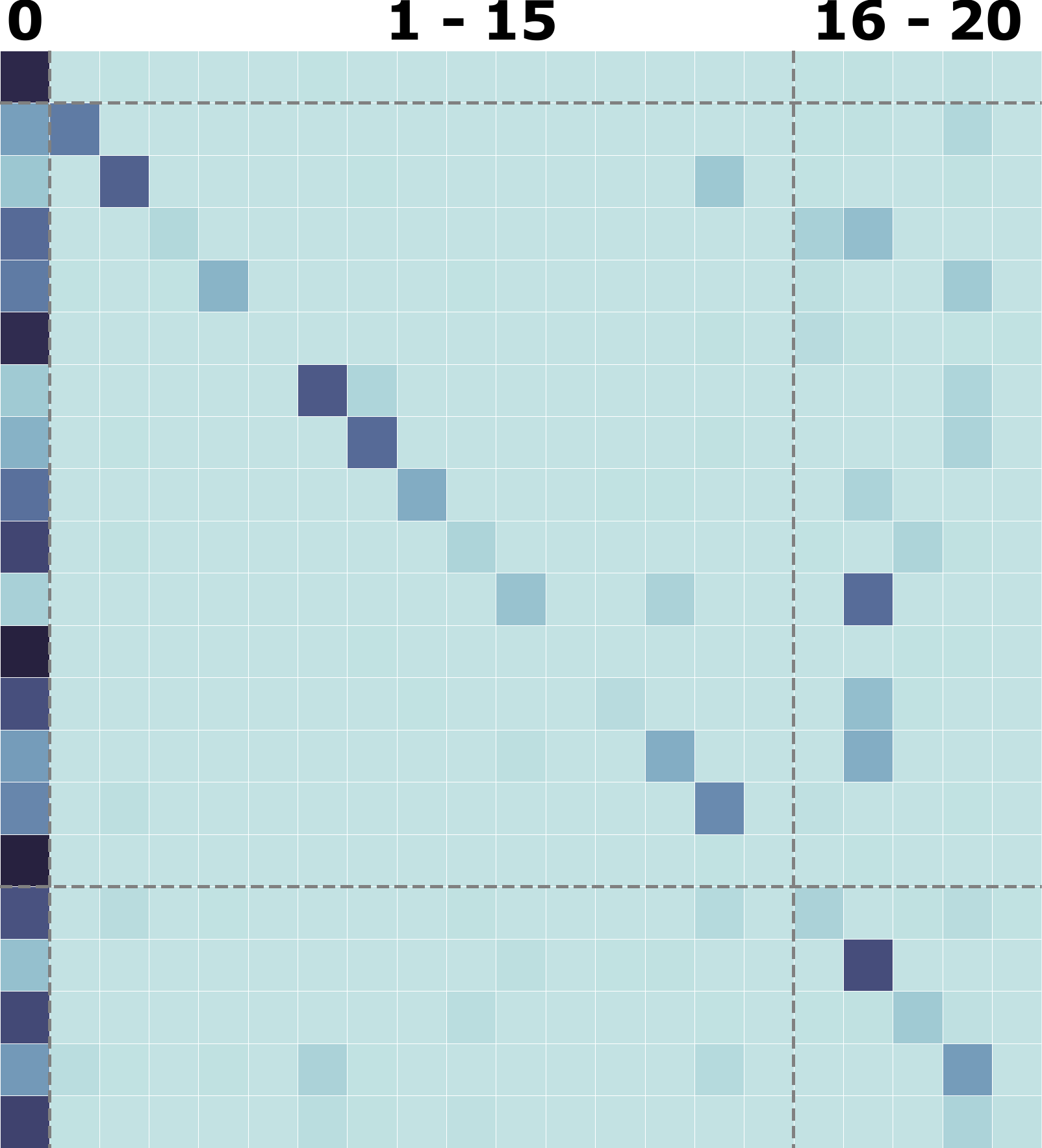}
        \caption{overlapped}
    \end{subfigure}    
    &  
        \begin{subfigure}[t]{0.28\textwidth}
        \includegraphics[width=\textwidth]{img/confmat/MAS_confmat.pdf}
        \caption{disjoint}
    \end{subfigure}
    & 
    \begin{subfigure}[t]{0.28\textwidth}
        \includegraphics[width=\textwidth]{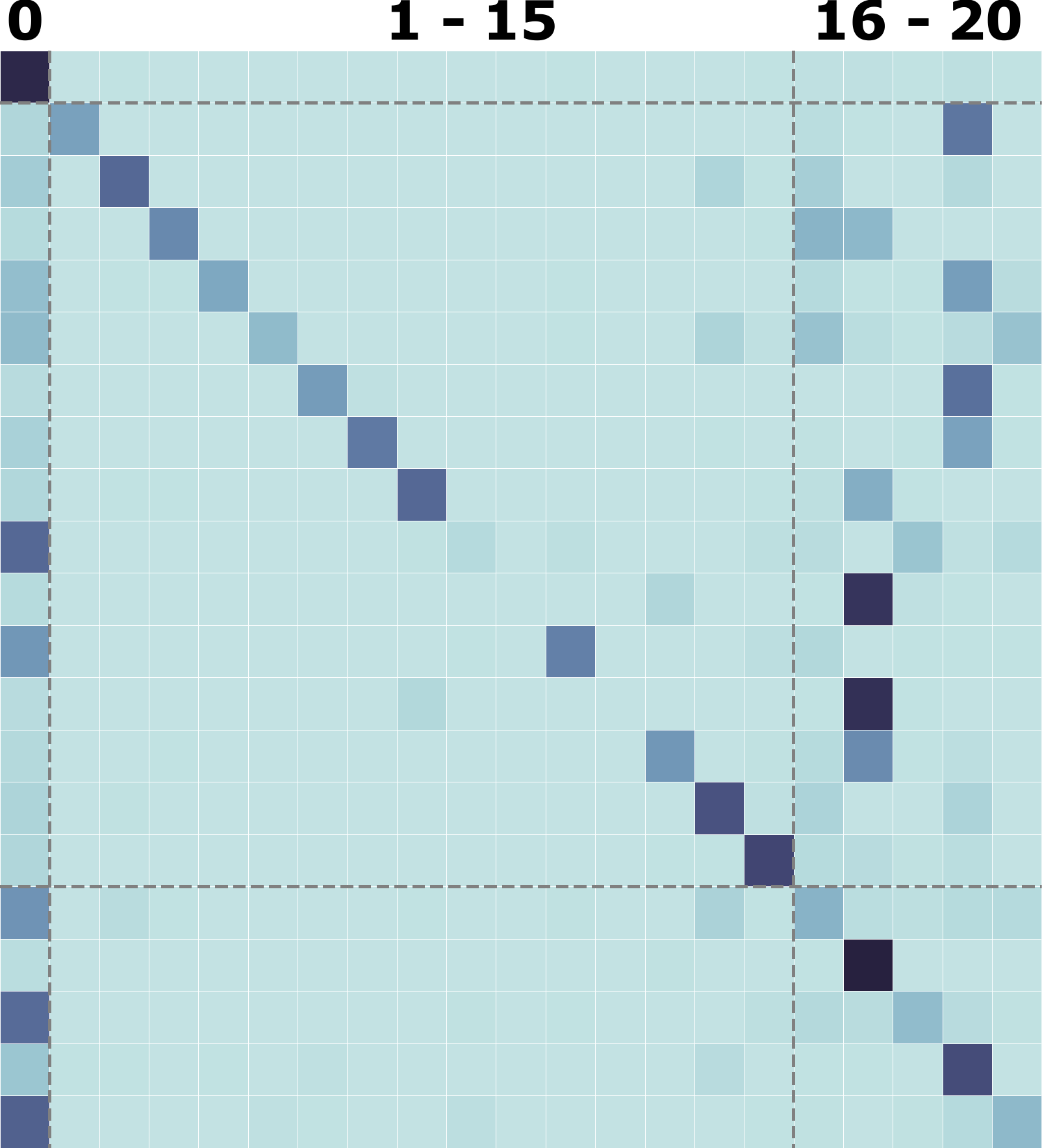}
        \caption{full-disjoint}
    \end{subfigure}
\end{tabular}%

\begin{tabular}{ccc}
\multicolumn{3}{c}{\textbf{EWC}} \\
    \centering
    \begin{subfigure}[t]{0.28\textwidth}
        \includegraphics[width=\textwidth]{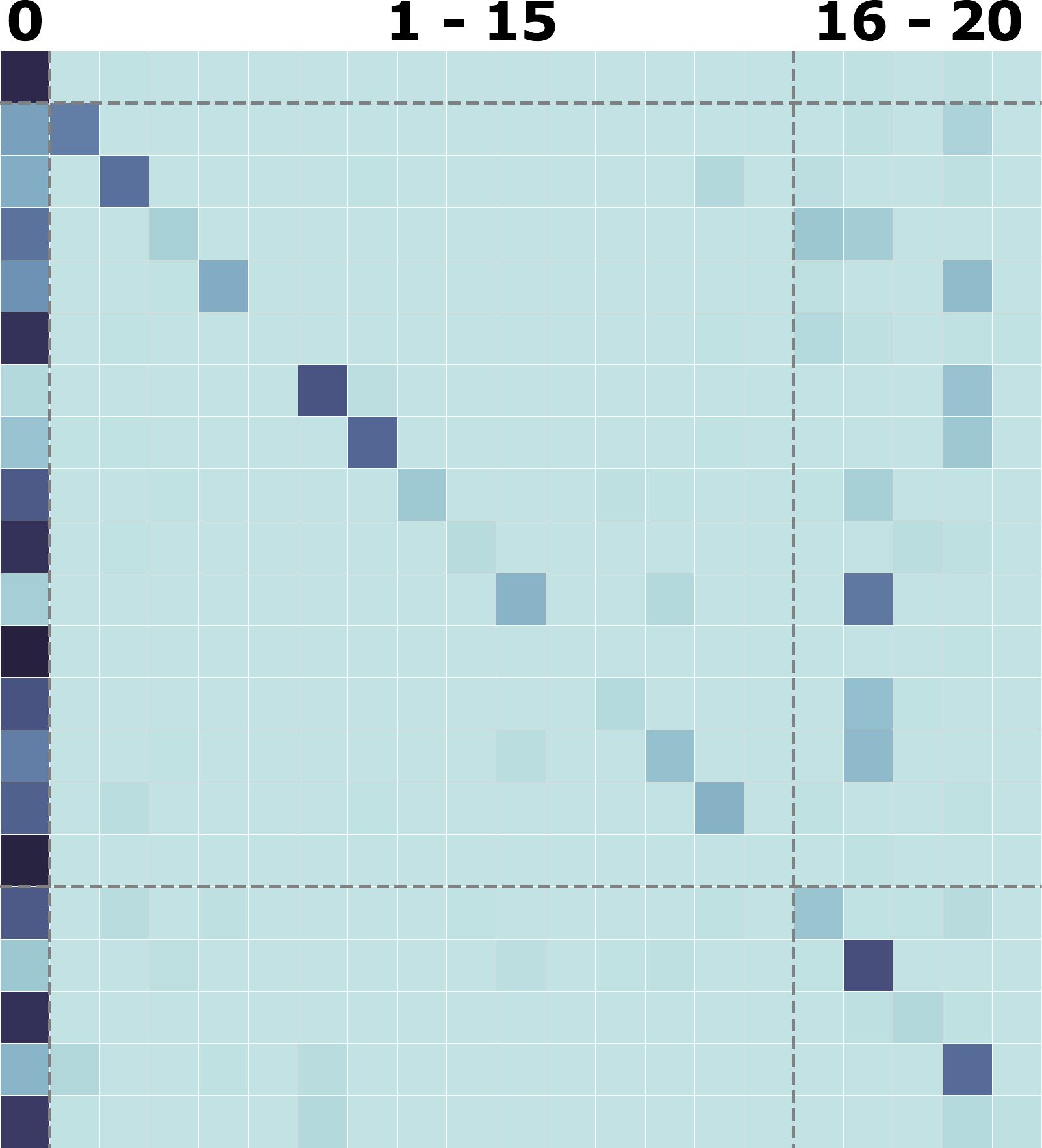}
        \caption{overlapped}
    \end{subfigure}    
    &  
        \begin{subfigure}[t]{0.28\textwidth}
        \includegraphics[width=\textwidth]{img/confmat/EWC_confmat.pdf}
        \caption{disjoint}

    \end{subfigure}
    & 
    \begin{subfigure}[t]{0.28\textwidth}
        \includegraphics[width=\textwidth]{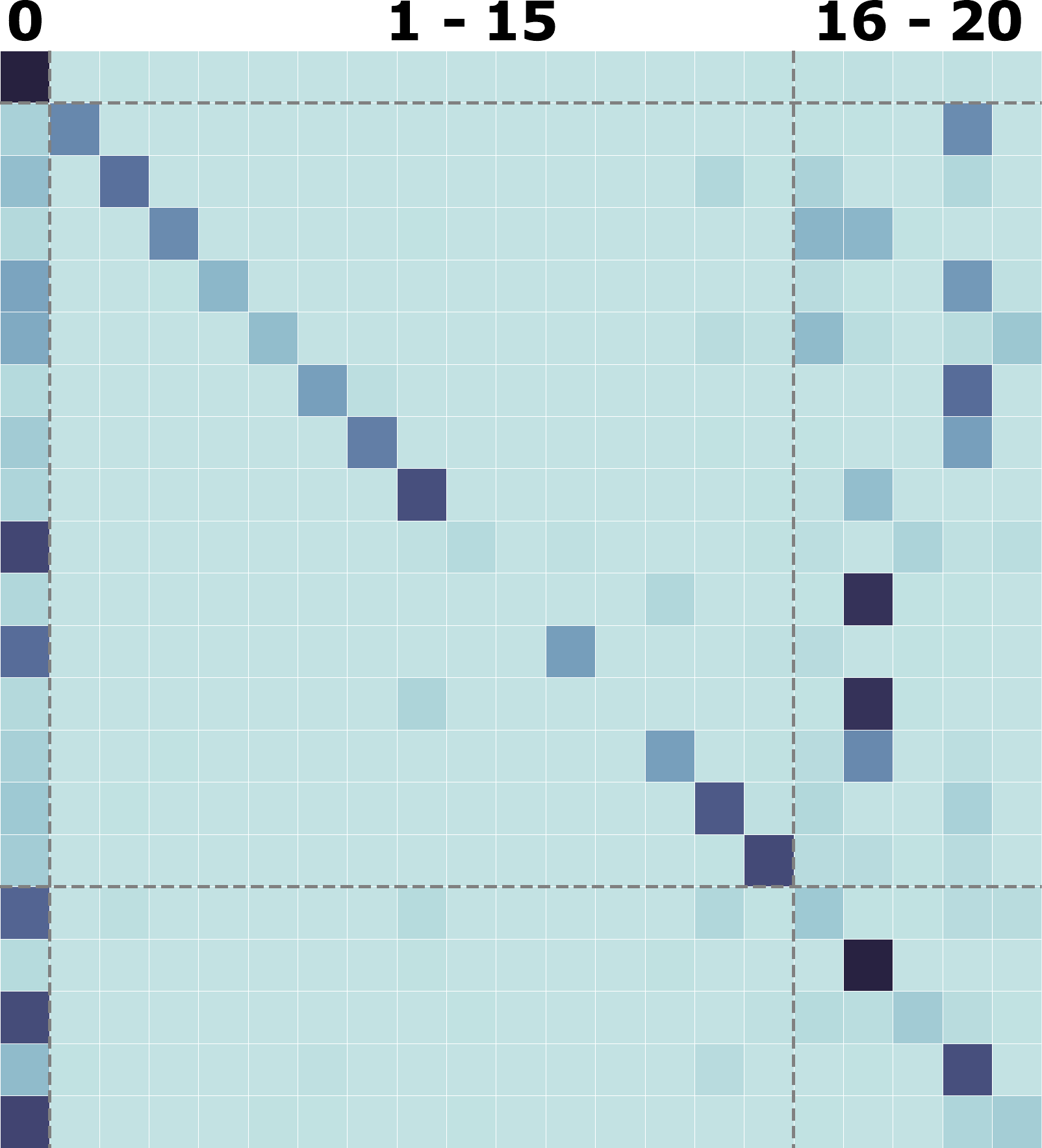}
        \caption{full-disjoint}
    \end{subfigure}
\end{tabular}%
\end{figure}
\begin{figure}[htbp]
\begin{tabular}{ccc}
\multicolumn{3}{c}{\textbf{LwF}} \\
    \centering
    \begin{subfigure}[t]{0.28\textwidth}
        \includegraphics[width=\textwidth]{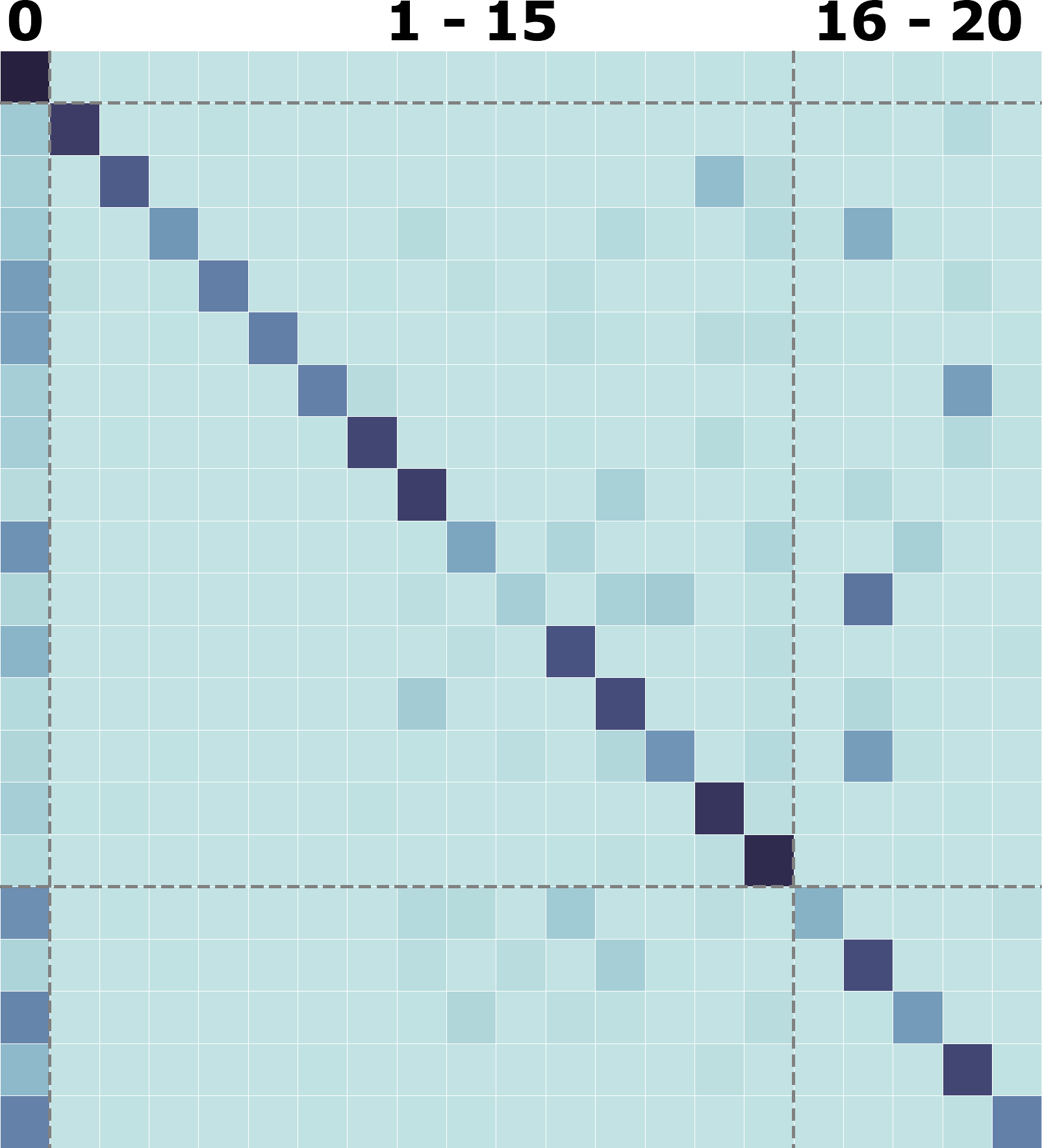}
        \caption{overlapped}
    \end{subfigure}    
    &  
        \begin{subfigure}[t]{0.28\textwidth}
        \includegraphics[width=\textwidth]{img/confmat/LwF_confmat.pdf}
        \caption{disjoint}
    \end{subfigure}
    
    & 
    \begin{subfigure}[t]{0.28\textwidth}
        \includegraphics[width=\textwidth]{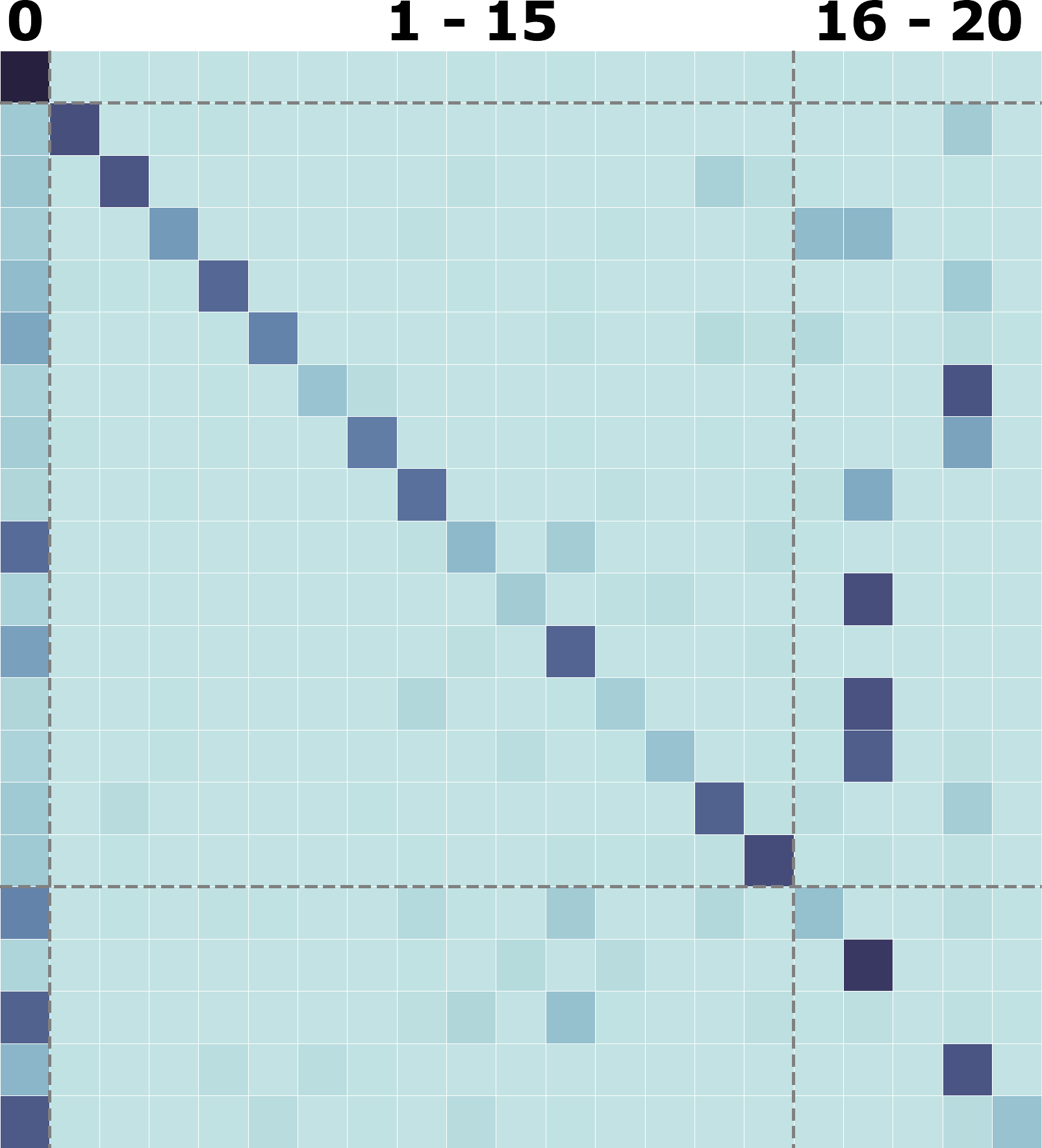}
        \caption{full-disjoint}
    \end{subfigure}
\end{tabular}%

\begin{tabular}{ccc}
\multicolumn{3}{c}{\textbf{Replay}} \\
    \centering
    \begin{subfigure}[t]{0.28\textwidth}
        \includegraphics[width=\textwidth]{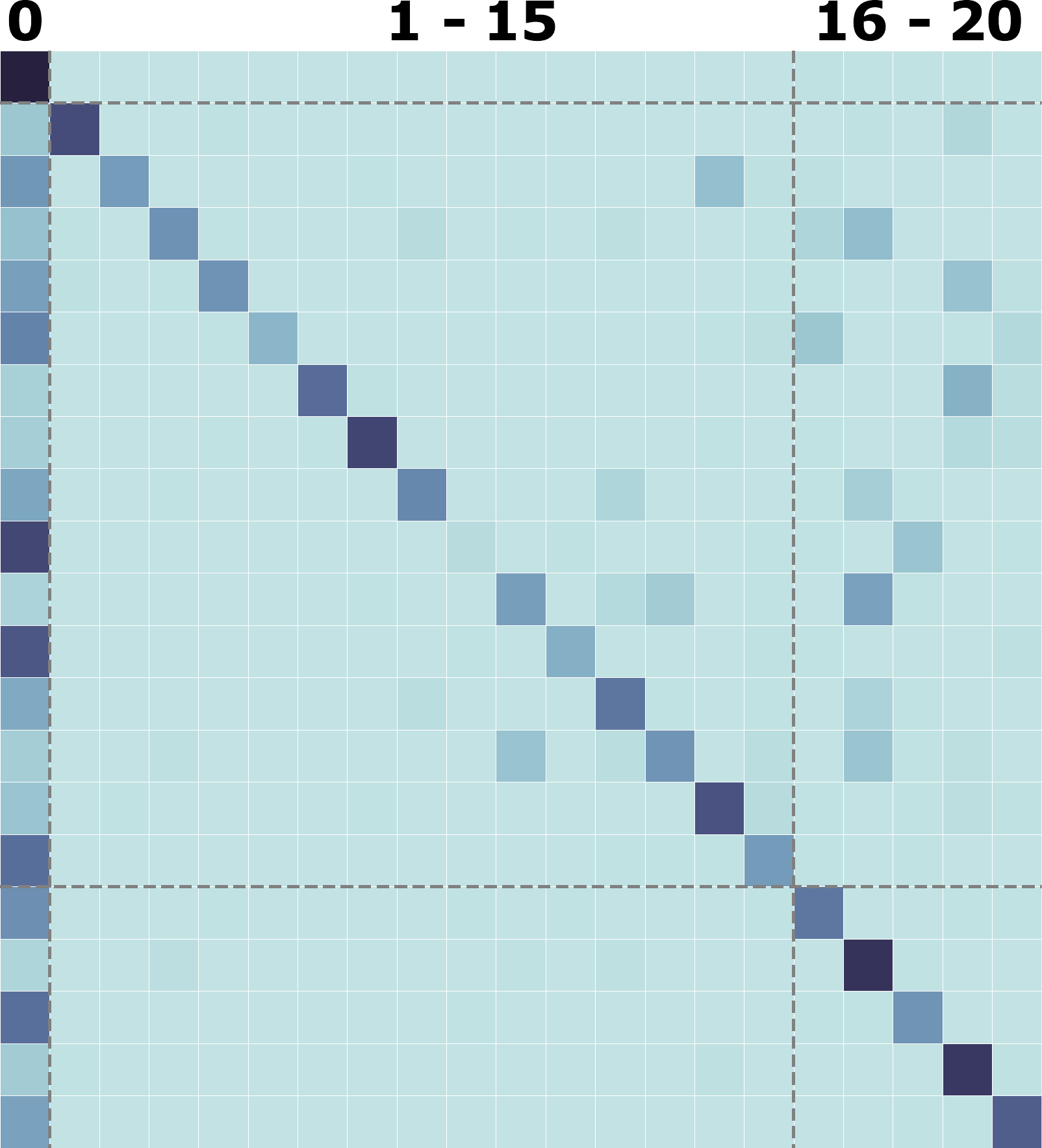}
        \caption{overlapped}
    \end{subfigure}    
    &  
        \begin{subfigure}[t]{0.28\textwidth}
        \includegraphics[width=\textwidth]{img/confmat/Replay_confmat.pdf}
        \caption{disjoint}
    \end{subfigure}
    
    & 
    \begin{subfigure}[t]{0.28\textwidth}
        \includegraphics[width=\textwidth]{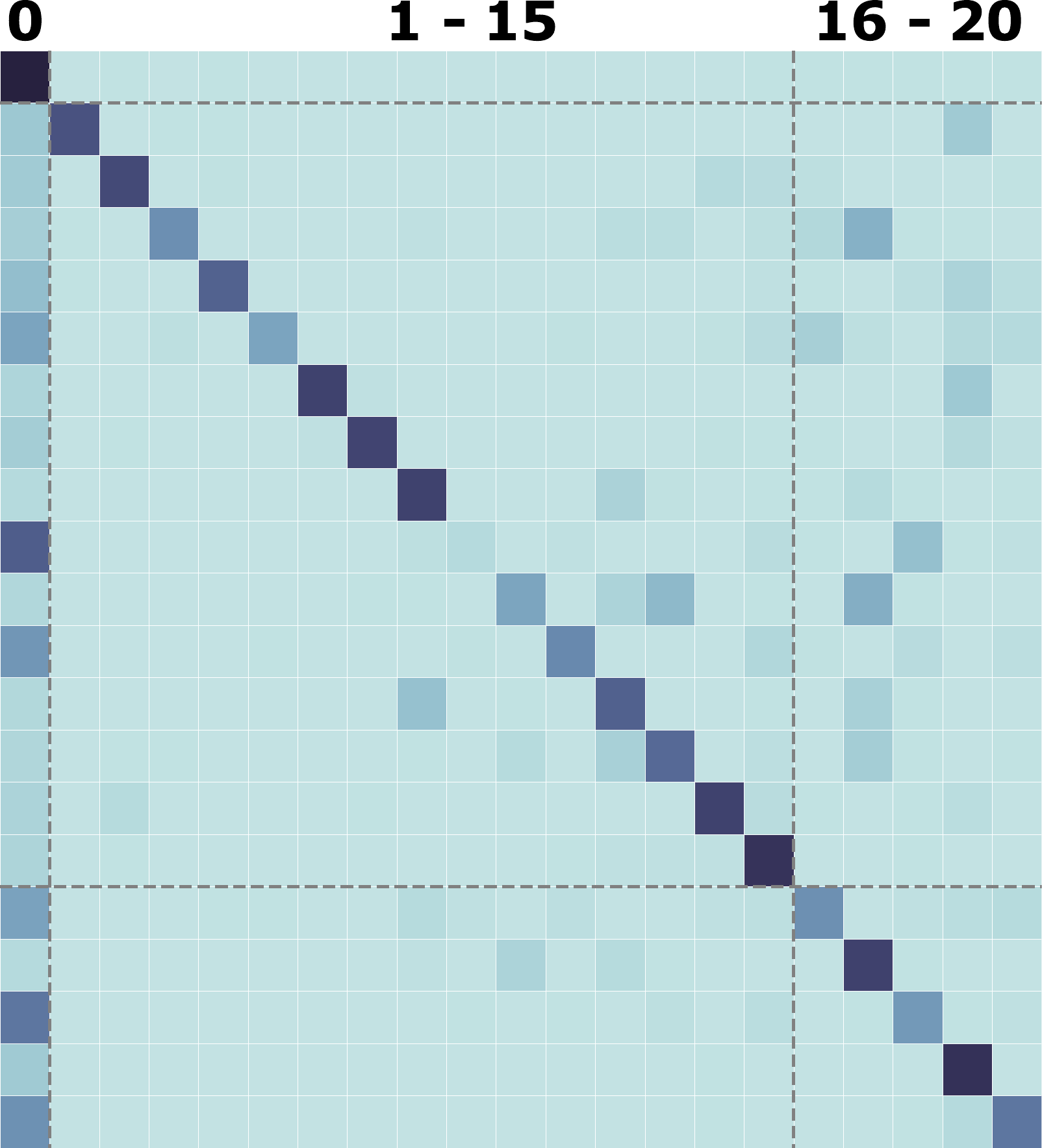}
        \caption{full-disjoint}
    \end{subfigure}
\end{tabular}%
\end{figure}

\end{document}